\documentclass[cvprfinal]{cvpr} 

\usepackage{times}
\usepackage{epsfig}
\usepackage{graphicx}
\usepackage{amsmath}
\usepackage{amssymb}
\usepackage{booktabs}
\usepackage{comment}
\usepackage{xcolor}
\usepackage[numbers,sort]{natbib}
\usepackage{multirow}
\usepackage{algorithm}
\usepackage{algorithmic}
\usepackage{dsfont}
\usepackage{bm}  
\usepackage{xspace}
\usepackage{subfigure}
\usepackage{colortbl}
\usepackage{enumitem}
\usepackage{placeins}
\usepackage{footnote}

\usepackage[utf8]{inputenc}
\usepackage{amsfonts}    
\usepackage{nicefrac}       
\usepackage{mathtools}
\usepackage{tabularx}
\usepackage{adjustbox}
\usepackage{threeparttable}
\usepackage{etoolbox}
\usepackage{lipsum}
\usepackage{cuted}
\usepackage{capt-of}

\usepackage{color}
\usepackage{kotex}

\usepackage{graphics}
\usepackage{siunitx}
\usepackage{blindtext}
\usepackage[normalem]{ulem}
\usepackage{float}
\usepackage{array}
\usepackage{tabulary}	
\usepackage{url}

\include{math_commands}
\definecolor{Gray}{gray}{0.9}

\newcommand{\eqnsm}[2]{\begin{equation}\label{eq:#1}#2\end{equation}}

\newlength\secmargin
\newlength\paramargin
\newlength\abovetabcapmargin
\newlength\belowtabcapmargin
\newlength\abovefigcapmargin
\newlength\belowfigcapmargin
\makeatletter\renewcommand\paragraph{\@startsection{paragraph}{4}{\z@}
  {.5em \@plus1ex \@minus.2ex}{-.5em}{\normalfont\normalsize\bfseries}}\makeatother

\usepackage[pagebackref=true,breaklinks=true,colorlinks,citecolor=citecolor,bookmarks=false]{hyperref}

\makeatletter
\newcommand{\thickhline}{%
    \noalign {\ifnum 0=`}\fi \hrule height 1pt
    \futurelet \reserved@a \@xhline
}

\DeclareRobustCommand\onedot{\futurelet\@let@token\@onedot}
\def\@onedot{\ifx\@let@token.\else.\null\fi\xspace}

\def\eg{\emph{e.g}\onedot} 
\def\ie{\emph{i.e}\onedot}

\usepackage{pifont}
\newcommand{\cmark}{\ding{51}}

\newcommand*\rot{\rotatebox{90}}

\newcommand{\argmin}{\mathop{\rm argmin}\limits}

\newcommand{\eqnref}[1]{Eq.~(\ref{#1})}

\newcommand{\enm}{Explore-And-Match\xspace}
\newcommand{\lvtr}{\textsc{Lvtr}\xspace}
\newcommand{\pb}{proposal-based\xspace}
\newcommand{\pf}{proposal-free\xspace}
\newcommand{\tll}{temporal localization loss\xspace}
\newcommand{\sgl}{set guidance loss\xspace}

\newcommand{\indic}[1]{\mathds{1}_{\{#1\}}}
\newcommand{\noobject}{\varnothing}
\renewcommand{\Sigma}{\mathfrak{S}}

\definecolor{citecolor}{HTML}{0071bc}
\definecolor{linkcolor}{HTML}{ED1C24}
\definecolor{urlcolor}{rgb}{0.2, 0.7, 0.1}
\definecolor{scorered}{HTML}{e4485a}
\definecolor{scoreblue}{HTML}{4a7ee8}
\definecolor{scoregreen}{HTML}{80ba0e}

\definecolor{purple0}{HTML}{e9e9f3}
\definecolor{purple}{HTML}{dcdaed}
\definecolor{purple1}{HTML}{bab6da}

\def\cvprPaperID{*****} 
\def\confName{CVPR}
\def\confYear{2022}

\begin{document}

\title{Explore-And-Match: Bridging Proposal-Based and Proposal-Free With Transformer for Sentence Grounding in Videos}

\author{Sangmin~Woo \quad
Jinyoung~Park \quad
Inyong~Koo \quad
Sumin~Lee \quad
Minki~Jeong \quad
Changick~Kim\\
Korea Advanced Institute of Science and Technology (KAIST)\\
{\tt\small \{smwoo95, jinyoungpark, iykoo010, suminlee94, rhm033, changick\}@kaist.ac.kr}
}
\maketitle

\begin{abstract}
Natural Language Video Grounding (NLVG) aims to localize time segments in an untrimmed video according to sentence queries.
In this work, we present a new paradigm named \enm for NLVG that seamlessly unifies the strengths of two streams of NLVG methods: \pf and \pb; the former \textit{explores} the search space to find time segments directly, and the latter \textit{matches} the predefined time segments with ground truths.
To achieve this, we formulate NLVG as a set prediction problem and design an end-to-end trainable Language Video Transformer (\lvtr) that can enjoy two favorable properties, which are rich contextualization power and parallel decoding.
We train \lvtr with two losses.
First, \textit{\tll} allows time segments of all queries to regress targets (explore).
Second, \textit{\sgl} couples every query with their respective target (match).
To our surprise, we found that training schedule shows divide-and-conquer-like pattern: time segments are first diversified regardless of the target, then coupled with each target, and fine-tuned to the target again.
Moreover, \lvtr is highly efficient and effective: it infers faster than previous baselines (by 2$\times$ or more) and sets competitive results on two NLVG benchmarks (ActivityCaptions and Charades-STA).
\end{abstract}

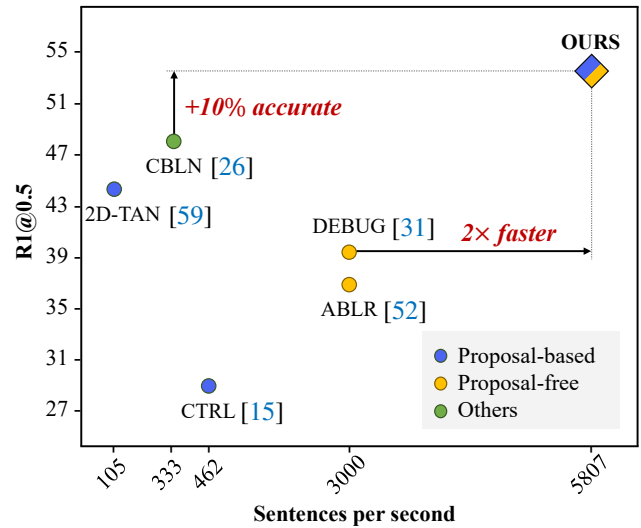
\begin{figure}[t]
    \centering
    \hspace{-3em}
    \def\svgscale{0.5}
    \def\svgwidth{\linewidth}
\begingroup%
  \makeatletter%
  \providecommand\color[2][]{%
    \errmessage{(Inkscape) Color is used for the text in Inkscape, but the package 'color.sty' is not loaded}%
    \renewcommand\color[2][]{}%
  }%
  \providecommand\transparent[1]{%
    \errmessage{(Inkscape) Transparency is used (non-zero) for the text in Inkscape, but the package 'transparent.sty' is not loaded}%
    \renewcommand\transparent[1]{}%
  }%
  \providecommand\rotatebox[2]{#2}%
  \newcommand*\fsize{\dimexpr\f@size pt\relax}%
  \newcommand*\lineheight[1]{\fontsize{\fsize}{#1\fsize}\selectfont}%
  \ifx\svgwidth\undefined%
    \setlength{\unitlength}{503.117188bp}%
    \ifx\svgscale\undefined%
      \relax%
    \else%
      \setlength{\unitlength}{\unitlength * \real{\svgscale}}%
    \fi%
  \else%
    \setlength{\unitlength}{\svgwidth}%
  \fi%
  \global\let\svgwidth\undefined%
  \global\let\svgscale\undefined%
  \makeatother%
  \begin{picture}(1,0.82769142)%
    \lineheight{1}%
    \setlength\tabcolsep{0pt}%
    \put(0,0){\includegraphics[width=\unitlength,page=1]{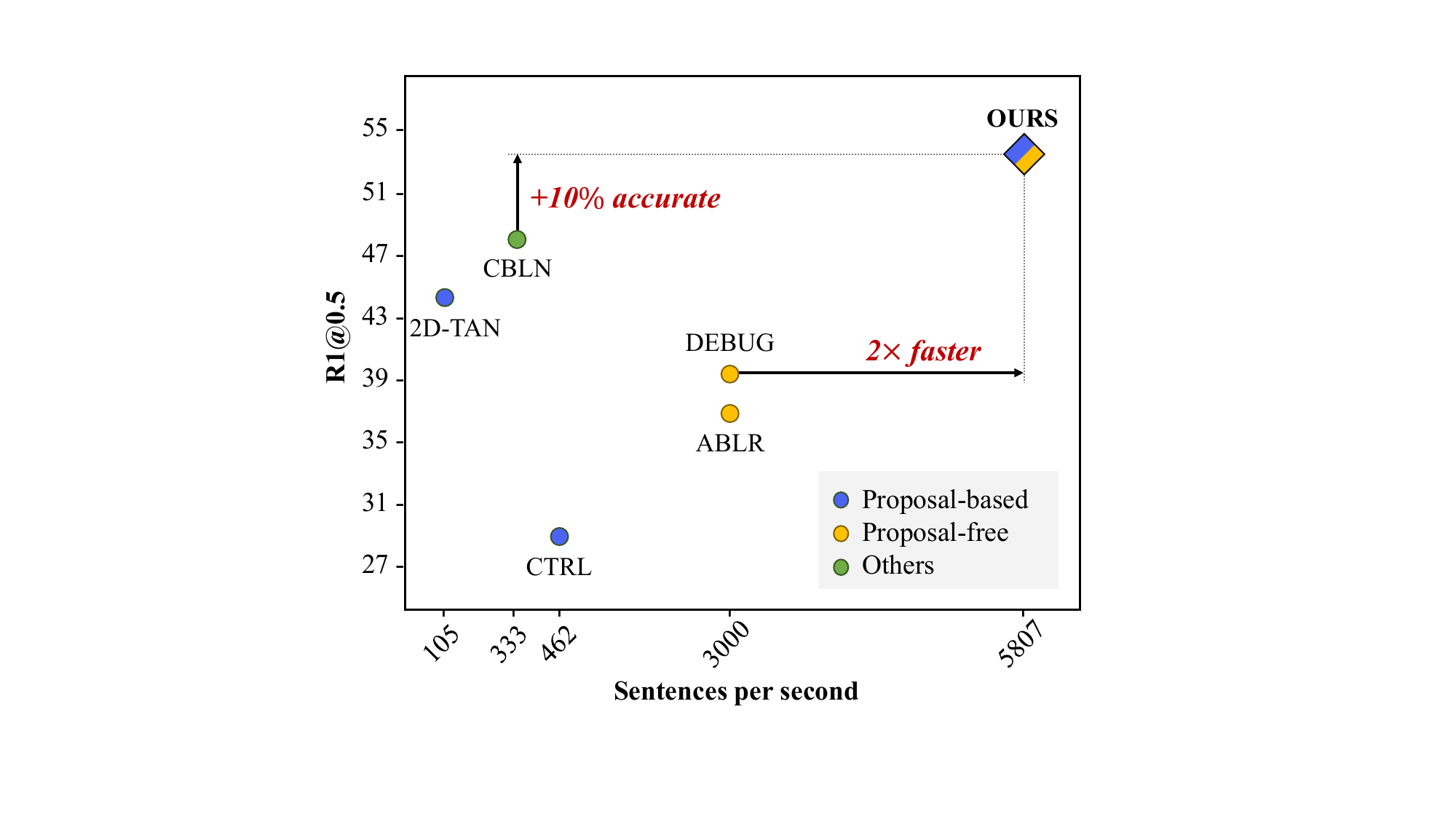}}%
    \put(0.31315685,0.56092568){\color[rgb]{0,0,0}\makebox(0,0)[lt]{\lineheight{1.25}\smash{\begin{tabular}[t]{l}\cite{liu2021context}\end{tabular}}}}%
    \put(0.24656739,0.48316999){\color[rgb]{0,0,0}\makebox(0,0)[lt]{\lineheight{1.25}\smash{\begin{tabular}[t]{l}\cite{zhang2020learning}\end{tabular}}}}%
    \put(0.60260659,0.46373106){\color[rgb]{0,0,0}\makebox(0,0)[lt]{\lineheight{1.25}\smash{\begin{tabular}[t]{l}\cite{lu2019debug}\end{tabular}}}}%
    \put(0.58930627,0.33226256){\color[rgb]{0,0,0}\makebox(0,0)[lt]{\lineheight{1.25}\smash{\begin{tabular}[t]{l}\cite{yuan2019semantic}\end{tabular}}}}%
    \put(0.36115474,0.17061273){\color[rgb]{0,0,0}\makebox(0,0)[lt]{\lineheight{1.25}\smash{\begin{tabular}[t]{l}\cite{gao2017tall}\end{tabular}}}}%
  \end{picture}%
\endgroup%

    \normalsize
    \caption{
        \lvtr achieves \textbf{10\% of performance gain} for the R1@0.5 metric while being \textbf{2$\times$ faster} than strong baselines on the ActivityCaptions dataset. The average inference speed is measured by the number of localized sentences per second.
    }
    \label{fig:teaser}
    \vspace{\belowfigcapmargin}
\end{figure}

\begin{figure*}[t!]
    \centering
    \includegraphics[width=1.0\linewidth]{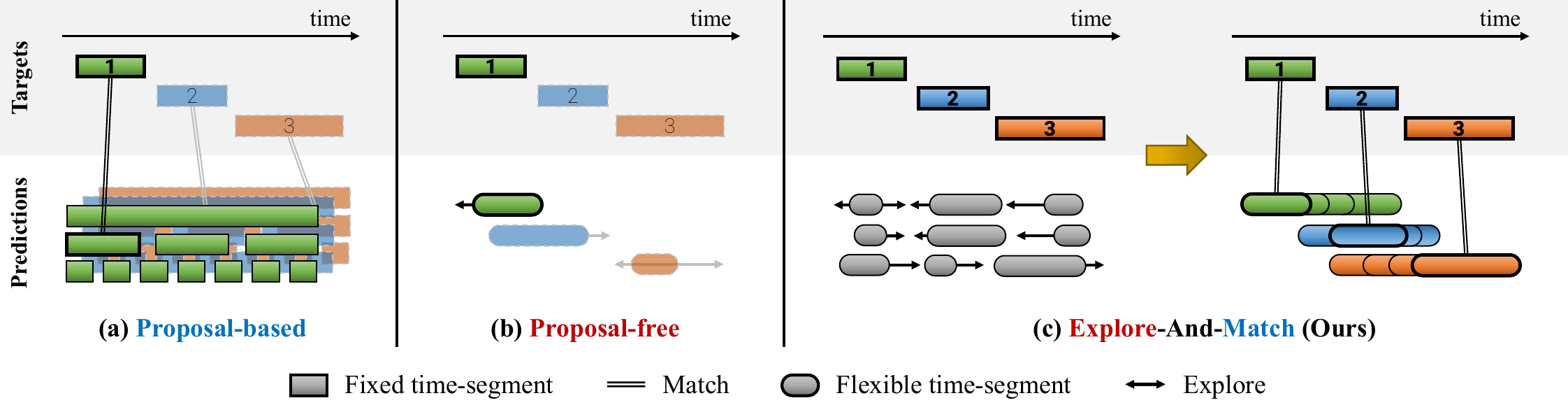}
    \caption{
        (a) \textbf{Proposal-free} methods directly regress start and end timestamps. (b) \textbf{Proposal-based} methods exhaustively match all predefined fixed-size proposals with ground truths. (c) Our \textbf{\enm} paradigm unifies two methods and instead makes flexible time segment proposals.
        Our method starts with randomly initialized proposals, \textit{explores} time space, and then \textit{matches} the corresponding target.
        By design, our \lvtr can predict multiple targets simultaneously, while earlier approaches could only predict single target at a time.
    }
    \label{fig:comparison}
    \vspace{\belowfigcapmargin}
\end{figure*}

\section{Introduction}
\label{sec:intro}

The explosion of video data brought on by the growth of the internet poses challenges to effective video search.
In order to accomplish successful video search, much effort has been put into language query-based video retrieval~\cite{miech2019howto100m,xu2019multilevel,gabeur2020multi,yang2020tree,chen2020fine}.
While text-video retrieval aims to match a trimmed video clip to the language query, NLVG aims to find accurate time segments relevant to the language queries in an untrimmed video.
It can be helpful especially when one wants to find a specific scene in a long video, such as a movie.
The majority of existing methods for NLVG can be categorized into two families: 1) \pb methods~\cite{anne2017localizing,gao2017tall,chen2018temporally,liu2018cross,hendricks2018localizing,zhang2019man,zhang2019cross,yuan2019semantic,ge2019mac,xu2019multilevel,wang2020temporally,zhang2020learning,qu2020fine,liu2020jointly,wang2021structured,zhang2021multi}, which generate a bunch of proposals in advance and select the best match with target segments, and 2) \pf methods~\cite{yuan2019find,ghosh2019excl,lu2019debug,chen2020rethinking,chen2020learning,mun2020local,wang2020dual,wang2020temporally,zeng2020dense,chen2020hierarchical,rodriguez2020proposal,zhang2020span,cao2021pursuit}, which estimate start and end timestamps aligned to the given description directly.
The \pb approaches generally show strong performance at the trade-off of the prohibitive cost of proposal generation.
They contradict the end-to-end philosophy, and their performances are significantly influenced by hand-designed pre-processing or post-processing steps such as dense proposal generation~\cite{chen2020look, xiao2021boundary} or non-maximum suppression~\cite{yuan2020semantic, wang2020temporally, soldan2021vlg} to abandon near-duplicate predictions.
On the other hand, the \pf approaches are much more efficient, but involve difficulties in optimization since the search space for segment prediction is too large.

In this work, we present a new NLVG paradigm named \textit{\enm} that combines the strengths of the two mainstream approaches by formulating NLVG as a direct set prediction problem.
Our method keeps the use of proposals while flexibly predicting time segments.
Also, it avoids time-consuming pre-processing and post-processing via a direct set prediction.
A conceptual comparison of our approach with two previous approaches is shown in~\figref{fig:comparison}.
To solve NLVG as a set prediction problem, we design an end-to-end trainable model called \lvtr based on the transformer encoder-decoder architecture~\cite{vaswani2017attention}.
The primary ingredients of \lvtr are bipartite set matching and parallel decoding with a small set of learnable proposals~\footnote{We refer to trainable positional encodings as \textit{learnable proposals} that are transformed into time segments by the transformer decoder.}.
To train all learnable proposals in parallel, we adopt the Hungarian algorithm~\cite{kuhn1955hungarian} to find the optimal bipartite matching (\ie, paired in a way that minimizes the matching cost) between ground truths and predictions.
This guarantees that each target has a unique match during training.
The self-attention mechanism of the transformer enables all elements in an input sequence to interact with one another, making transformer architecture particularly suitable for certain constraints of set prediction, such as suppressing duplicate predictions.
By design, \lvtr allows us to forgo the use of manually-designed components (\eg, temporal anchors, windows) that encode prior knowledge into the NLVG pipelines.
Furthermore, learnable proposals can interact with visual-linguistic representations as well as themselves to directly output the final time segment predictions in a single run.

Under the \textit{\enm} scheme, the overall training schedule is governed by \tll and \sgl, where the \tll is responsible for generating accurate time segments, and \sgl is responsible for matching predictions with their respective targets (\ie, making target-specific predictions).
To match the learnable proposals with their targets, we first divide the learnable proposals by the number of query sentences into several subsets, then \sgl progressively forces each subset of proposals to match its corresponding query.
In the early stages of training, the \tll holds the major term in set matching than \sgl, which means that it is more of a priority for each subset to somehow approximate the time segment regardless of the target than to predict the corresponding target.
Therefore, at first, random subsets learn to reduce \tll in a target-agnostic manner.
Then, once the \sgl becomes more dominant than \tll, the subsets begin to predict each designated target.
Finally, all learnable proposals learn to accurately align their their respective target segments.
learnable proposals diversify as they \textit{explore} time space, and then \textit{match} their respective targets.
While the training \lvtr under the \textit{\enm} scheme conforms to the end-to-end basis, it spontaneously divides and conquers the whole process rather than optimizing all the objectives simultaneously.
We show the empirical evidence of the \enm phenomenon (see~\figref{fig:explore_and_match}) and confirm that this simple strategy is remarkably effective (see~\figref{fig:teaser}).

We evaluate \lvtr trained under \enm scheme on two challenging NLVG benchmarks --- \textit{ActivityCaptions}~\cite{caba2015activitynet, krishna2017dense} and \textit{Chrades-STA}~\cite{gao2017tall} --- against the recent works.
Our \lvtr achieves new state-of-the-art results on two benchmarks, even without human priors such as knowledge of time segment distribution.
Lastly, we confirm the effectiveness of our approach by conducting extensive ablation studies and analyses.
To summarize, our contributions are three-fold:
\begin{itemize}[nosep]
    \item We present ``\enm", a new NLVG paradigm that unifies the strengths of \pb and \pf methods by combining our new \sgl with \tll.
    \item We propose an end-to-end trainable model, \lvtr, which models the NLVG as a set prediction problem. By design, our \lvtr can predict multiple sentence queries at once. Moreover, this formulation streamlines the overall pipeline by removing the use of several heuristics.
    \item Comprehensive experiments and extensive ablation studies demonstrate the effectiveness of \lvtr. Last but not least, \lvtr establishes a new state-of-the-art on two NLVG benchmarks while accelerating inference time (by 2$\times$ or more) than previous methods.
\end{itemize}

\section{Related Work}
\label{sec:rel_work}
\vspace{\paramargin}\paragraph{Video Grounding.}
The origin of NLVG traces back to the temporal activity localization~\cite{shou2016temporal}, which attempts to locate the start and end timestamps of actions and identify its labels in an untrimmed video.
Likewise, NLVG aims to retrieve the corresponding time segments, but it is grounded on language queries rather than a fixed set of action labels.
Pioneering NLVG works~\cite{anne2017localizing, gao2017tall} define the task and provide benchmark datasets. Since then, numerous efforts have been made to push the boundaries of NLVG.
Early works follow the \pb pipeline~\cite{anne2017localizing,gao2017tall,chen2018temporally,liu2018cross,hendricks2018localizing,zhang2019man,zhang2019cross,yuan2019semantic,ge2019mac,xu2019multilevel,wang2020temporally,zhang2020learning,qu2020fine,liu2020jointly,wang2021structured,zhang2021multi}, which segments a huge number of candidates at regular intervals on different scales, and then ranks them using an evaluation network.
While \pb approaches provide reliable results, they are highly dependent on proposal quality and suffer from the prohibitive cost of creating proposals, as well as the computationally inefficient comparison of all proposal-target pairings.
Another line of works is the \pf approaches~\cite{yuan2019find,ghosh2019excl,lu2019debug,chen2020rethinking,chen2020learning,mun2020local,wang2020dual,wang2020temporally,zeng2020dense,chen2020hierarchical,rodriguez2020proposal,zhang2020span,cao2021pursuit}, which tries to regress the time segments directly. They are more flexible than \pb approaches in terms of granularity.
However, its accuracy generally lags behind that of its counterpart.
To summarize, the former tries to \textit{match} the predefined proposals with ground truth, while the latter \textit{explores} the whole search space to find time segments directly.

This work aims to integrate two streams of NLVG methods into a single paradigm named \enm, by formulating NLVG as a direct set prediction problem.
Our method generates flexible time segments like \pf approaches while preserving the concept of \pb approaches that use positive and negative proposals at the same time.


\vspace{\paramargin}\paragraph{Transformers.}
A transformer~\cite{vaswani2017attention} is a universal sequence processor with an attention-based encoder-decoder architecture.
The self-attention mechanism captures both long-range interactions in a single context, and the encoder-decoder attention accounts for token correspondences across multi modalities.
Due to the tremendous promise of the attention mechanism, transformers have recently demonstrated their potential in various computer vision tasks: object detection~\cite{carion2020end}, video instance segmenation~\cite{wang2021end}, panoptic segmentation~\cite{wang2021max}, human pose and mesh reconstruction~\cite{lin2021end}, lane shape prediction~\cite{liu2021end}, and human object interaction~\cite{zou2021end}.

Among them, it is worth noting that the Detection Transformer (DETR)~\cite{carion2020end}, the first transformer-based end-to-end object detector, achieved very competitive performance despite its simple design.
DETR successfully removes many hand-crafted components from the object detection pipeline by using powerful relation modeling capabilities of transformer.
The principal component of DETR is bipartite matching, notably the Hungarian algorithm~\cite{kuhn1955hungarian}, which generates a set of unique bounding boxes.
This saves a lot of post-processing time by removing non-maximum suppression from the pipeline.
Also, DETR infers a set of predictions in parallel with a single iteration through the decoder.

Inspired by the recent successes of transformers, we propose a novel NLVG model named Language Video Transformer (\lvtr) based on the transformer architecture.
The attention mechanism of the transformer allows every element of the input sequence to attend to each other while utilizing rich contextualization.
This architectural strength makes the transformer particularly suitable for our NLVG formulation, a direct set matching problem.
We note that final time segment predictions are directly generated in an end-to-end manner.

\begin{figure*}[t!]
    \centering
    \includegraphics[width=1.0\linewidth]{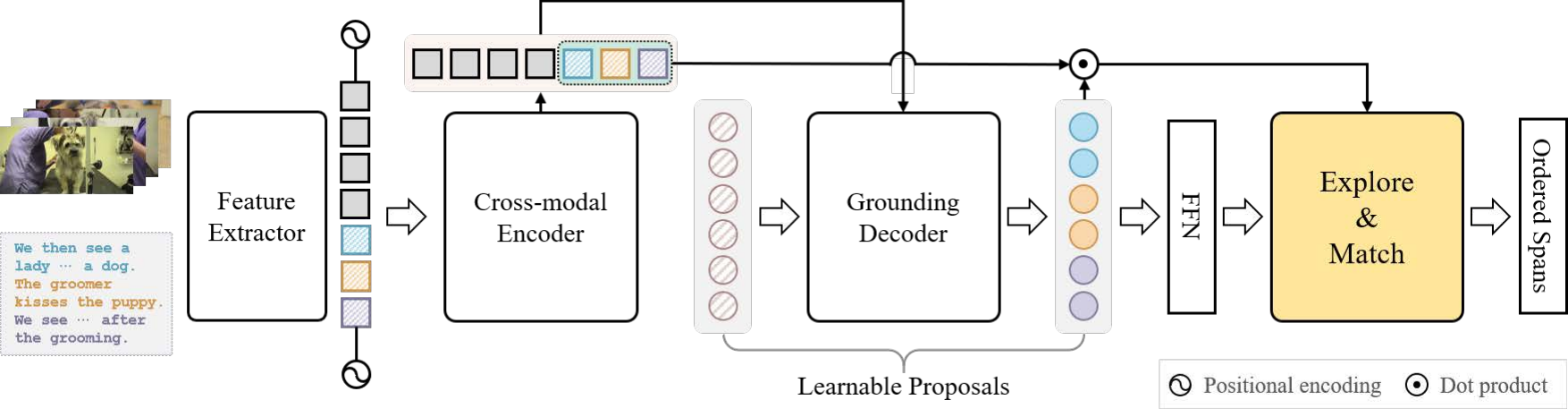}
    \caption{
        \textbf{Overview of \lvtr.}
        From the \textbf{feature extractor}, we first obtain video and text features and supplement them with positional encoding.
        The \textbf{encoder} takes as input a sequence of concatenated video-text features.
        The \textbf{decoder} is fed with a fixed number of \textit{learnable proposals}, which in turn attend to themselves and the encoder output, generating contextualized outputs.
        These outputs are then used to predict time segments via a FFN and used to measure the correspondence (\eg, normalized similarity) with the textual outputs of the encoder using a dot product ($\odot$).
        Following that, the overall training process follow the \textit{\enm} scheme (more details are in~\figref{fig:explore_and_match}).
        The \lvtr is trained end-to-end, and it can directly output a set of ordered time segments in parallel. 
    }
    \label{fig:lvtr}
    \vspace{\belowfigcapmargin}
\end{figure*}

\section{Preliminary: Transformer}
Given that our model is built on the Transformer design, we will briefly discuss the general form of attention mechanisms, a key building block of Transformer.
The common practice~\cite{vaswani2017attention} is to use residual connections, dropout, and layer normalization.
The attention mechanism is described in depth in~\cite{vaswani2017attention}.

\vspace{\paramargin}\paragraph{Self-Attention.}
Self-Attention (SA) in the general {\bf qkv} form is a popular yet strong mechanism for neural systems.
We calculate a weighted sum over all values {\bf v} for each element in an input sequence \({\bf x} \in \mathbb{R}^{S \times D}\).
The attention weights \(A_{ij}\) are calculated by comparing two elements of the sequence to their respective query \({\bf q}_i\) and key \({\bf k}_j\) representations.
\begin{equation}
    [{\bf q, k, v}] = {\bf x}{{\bf W}_{qkv}} + {\bf p},
\end{equation}
\begin{equation}
    A = {\rm softmax}\left({\rm\bf qk}^T / \sqrt{D_h} \right),
\end{equation}
\begin{equation}
    {\rm SA}({\bf x}) = A{\bf v},
\end{equation}
where \({{\bf W}_{qkv}} \in \mathbb{R}^{D \times 3D_h}\) and \(A \in \mathbb{R}^{S \times S}\) are learnable weights. Since the Transformer is inherently permutation-invariant w.r.t input sequence, we add positional encoding \({\bf p} \in \mathbb{R}^{S \times D}\)~\cite{cordonnier2019relationship} to embedded sequence in practice.

\vspace{\paramargin}\paragraph{Multi-Head Self-Attention.}
Multi-Head Self-Attention (MHSA) is a simple extension of self-attention in which several self-attentions, dubbed "heads", are executed in parallel followed by a projection of their concatenated outputs.
To maintain the computed value and the number of parameters constant when changing \(k\), \(D_h\) is typically set to \(D/k\).
\begin{equation}
    {\rm MSHA({\bf x})} = [{\rm SA}_1({\bf x}); {\rm SA}_2({\bf x}); \cdots; {\rm SA}_k({\bf x});]{{\bf W}_{msha}},
\end{equation}
where [;] denotes concatenation on the channel axis and \({{\bf W}_{msha}} \in \mathbb{R}^{k\cdot D_h \times D}\) is learnable weight.

\section{Method}
We first define the NLVG task and propose our end-to-end trainable \lvtr.
Next, we describe our training losses and set matching strategy.
Finally, we present \enm, a novel paradigm that unifies \pb and \pf methods.

\subsection{Problem Formulation}
NLVG aims to localize a set of language-grounded time segments in an untrimmed video.
Since NLVG does not have a fixed set of sentence classes, the conventional classification approach is not applicable (\ie, taxonomy-free).
Therefore, the NLVG model should be able to infer the time segments while not being constrained by the predefined categories.
Formally, given a video $\mathcal{V}$ with a set of language queries $\mathcal{Q} = \{q_i\}^{K}_{i=1}$, we require a set of corresponding time segments.
\eqnsm{yi}{\{y_i\}_{i=1}^K = \{(t_i, q_i)\}_{i=1}^K \,,}
where $t_i = (t^s_i, t^e_i) \in [0, 1]$ defines the start and end timestamp normalized by the video length (\ie, time segment), and $K$ is the number of the queries. If $K=1$, the model only expects a single sentence as an input query, which is a conventional single-query setting.
In this setting, there is no need for prediction-query assignment since all the predictions of learnable proposals can be associated with only one target (\ie, $q_i$ can be omitted in~\eqnref{eq:yi}).
However, this limits the abundant interactions of the transformer with parallel decoding.
In order to account for beneficial semantic and temporal relationships between the time segments, we view NLVG as a direct set prediction problem.
In a multi-query setting, the model needs to specify which predictions are paired with which queries. Therefore, the grounding model should assign correct queries to the estimated time segments:
\eqnsm{hyi}{\{\hat y_i\}_{i=1}^N = \{(\hat t_i, \hat q_i)\}_{i=1}^N \,,}
where $\hat{t}$ and $\hat{q}$ denote the predicted time segments and queries, respectively. The number of predictions $N$ is substantially larger than the actual number of queries $K$ in the video.

\begin{figure}[t!]
    \centering
    \small
    \includegraphics[width=\columnwidth]{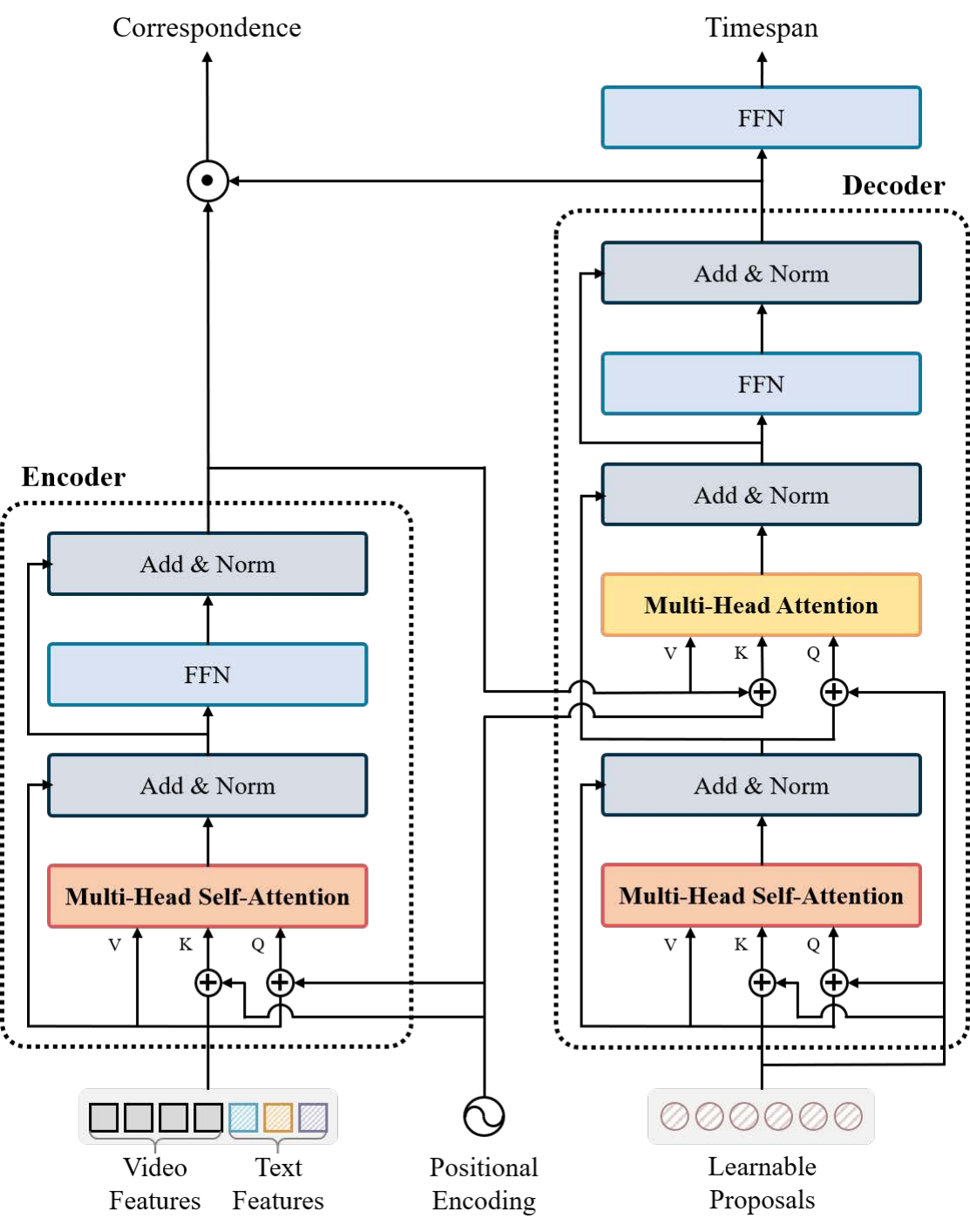}
    \caption{\textbf{Detailed \lvtr architecture.}}
    \label{fig:transformer}
    \vspace{\belowfigcapmargin}
\end{figure}

\subsection{\lvtr Architecture}
The overall pipeline of \lvtr is illustrated in~\figref{fig:lvtr}.
\lvtr contains three main components: 1) a feature extractor to obtain compact video and text representations, 2) a transformer encoder-decoder for contextualization and parallel decoding, and 3) a feed-forward network (FFN) that makes the final segment predictions.

The architectural details of the \lvtr are in~\figref{fig:transformer}.
The overall design is similar to that of original transformer encoder-decoder~\cite{vaswani2017attention}.
First, the transformer encoder processes video-text features, which are extracted from the backbone, added with temporal positional encoding~\footnote{We use a fixed absolute encoding to represent the temporal positions.} at each multi-head self-attention layer.
Next, the decoder receives learnable proposals and encoder memory and process them with multiple multi-head self-attention and encoder-decoder attention layers.
Finally, the output of decoder is used to generate the final set of predicted time segments, and also used to measure the correspondence between proposals and text queries.

\vspace{\paramargin}\paragraph{Feature Extractor.}
An input video $\mathcal{V} \in \mathbb{R}^{T_0 \times C_0 \times H_0 \times W_0}$ passes through the C3D~\cite{tran2015learning} (typical values we use are $T_0=16, C_0=3$  and $H_0=W_0=112$), and is transformed into a video feature $f_v \in \mathbb{R}^{T \times C \times H \times W}$ ($T=1, C=512$ and $H=W=4$). Since the input to transformer encoder should be in the form of sequence, we collapse the channel and spatial dimensions into a single dimension ($T \times CHW$). Then, we feed the output into a linear layer, which yields $T \times D$ dimensions.
On the other hand, input language queries $\mathcal{Q}$ break down into a set of word sequences, and then are converted into GloVe~\cite{pennington2014glove} embeddings.
A set of sentence representations $f_t \in \mathbb{R}^{K \times D}$ ($K \geq 1, D=512$) is obtained via a 2-layer bi-LSTM~\cite{hochreiter1997long}, followed by a linear layer.
The input sentences are batch-processed by applying zero-padding to have the same dimension $K$ as the largest number of sentences within the batch.
For a fair comparison, \lvtr is equipped with a conventional C3D+LSTM backbone, but it can be trained on top of any modern backbone (e.g., CLIP~\cite{radford2021learning}, ViT~\cite{dosovitskiy2020image}).

\vspace{\paramargin}\paragraph{Language Video Transformer.}
To begin, video and text features are obtained using their respective feature extractors.
We concatenate video-text features and pass them into the transformer encoder.
The transformer is unable to preserve the order of temporally arranged video features due to the permutation-invariant nature of the architecture. Therefore, we add fixed positional encodings to concatenated video-text features at every attention layer.
Each encoder layer has two sub-layers: a multi-head self-attention layer and a feed-forward network.
The key component of the encoder is self-attention, which relates different positions of a single sequence to compute an intra-representation of the sequence.
The decoder structure adds encoder-decoder attention in addition to the two sub-layers in the encoder.
The decoder takes a fixed-size set of $N$ inputs, which we refer to as \textit{learnable proposals}, and decodes them into a set of $N$ output embeddings.
All proposals collaboratively generate predictions in a set-wise manner with self-attention while accessing the whole video-text context with encoder-decoder attention.
The output embeddings from the decoder are fed into the prediction head, resulting in $N$ final time segment predictions.
The prediction head is a 2-layer perceptron with a two-dimensional output, which is set to predict start and end timestamps.
To match the proposals to corresponding sentences, we measure their correspondence with the normalized similarity of the decoder output and textual output of the encoder.
This is used to link each prediction to the query with the highest similarity.

\subsection{Explore-And-Match}
Considering that video includes multiple events over various periods, we view NLVG as a set prediction problem.
To solve a set prediction problem between predicted and ground truth time segments, we adopt a Hungarian matching algorithm~\cite{kuhn1955hungarian}.
We define our loss based on the set matching results.
Several training ingredients condense into \enm, a new paradigm that combines two streams of methods, \pb and \pf.

\vspace{\paramargin}\paragraph{NLVG as a set prediction.}
We search for one-to-one matching between the prediction set $\{\hat y_i\}_{i=1}^N$ and the ground truth set $\{y_i\}_{i=1}^K$ that optimally assigns predicted time segments to each ground truth.
We assume that the number of predictions $N$ is sufficiently larger than the number of queries $K$ in the video. Therefore, we consider the ground truth set $y$ as a set of size $N$ padded with $\noobject$ (no matching) for one-to-one matching.
We define a set of all permutations that consist of $N$ items as $\Sigma_N$. Among the set of permutations $\Sigma_N$, we seek an optimal permutation $\hat{\sigma} \in \Sigma_N$ that best assigns the predictions at the lowest cost:
\eqnsm{match_cost}{\hat{\sigma} = \argmin_{\sigma\in\Sigma_N} \sum_{i=1}^{N} \mathcal{C}_{match}(y_i, \hat y_{\sigma(i)}) \,,}
where $\mathcal{C}_{match}(y_i, \hat y_{\sigma(i)})$ is a pair-wise matching cost between ground truth $y_i$ and a prediction with index $\sigma(i)$. We detail the matching cost in~\eqnref{eq:match_cost_2}.

\vspace{\paramargin}\paragraph{Set guidance loss.}
By the permutation-invariant nature of the transformer, the prediction order cannot be determined.
This raises a question: how can we match the predictions with corresponding queries?
To address this problem, we introduce a \sgl that forces each prediction to be associated with a particular language query.
Given $K$ input queries, $N$ proposals are uniformly partitioned into $K$ subsets.
The proposals within the $j$th subset are trained to predict the $j$th query by \sgl.
Formally, we denote the probability that the prediction corresponds to the target query $q_i$ (\ie, softmaxed correspondence) as $\hat p_{\sigma(i)}(q_i)$ for the prediction with index $\sigma(i)$.
The \sgl is simply defined as a negative log-likelihood loss:
\eqnsm{guidance_loss}{\mathcal{L}_{sg}(q_i) = -\sum_i \log \hat p_{\sigma(i)}(q_i) \,.}
While all proposals collaboratively predict a set of time segments via parallel decoding, the \sgl leads proposals to predict target-specific time segments.

\vspace{\paramargin}\paragraph{Temporal localization loss.}
Our \tll is a linear combination of the $\ell_1$ loss and the generalized IoU (gIoU) loss~\cite{rezatofighi2019generalized}:
\eqnsm{localization_loss}{\mathcal{L}_{loc}(t_{i}, \hat t_{\sigma(i)}) = \lambda_{\rm L1}\mathcal{L}_{\rm L1}(t_{i}, \hat t_{\sigma(i)}) + \lambda_{\rm iou}\mathcal{L}_{iou}(t_{i}, \hat t_{\sigma(i)})\,,}
where $t_i$ is the ground truth time segment and $\hat t_{\sigma(i)}$ is the predicted time segment for the prediction with index $\sigma(i)$.
$\lambda_{\rm L1}, \lambda_{\rm iou} \in \mathbb{R}$ are balancing hyperparameters.
While two loss terms share the same objective, they have subtle differences.
The $\ell_1$ loss will have different scales for short and long time segments, even if relative errors are similar, whereas gIoU loss is robust to varying scales.
\eqnsm{l1_loss}{\mathcal{L}_{\rm L1}(t_{i}, \hat t_{\sigma(i)}) = ||t_{i}^{s}- \hat t_{\sigma(i)}^{s}||_1 + ||t_{i}^{e} - \hat t_{\sigma(i)}^{e}||_1\,,}

\begin{align}
\mathcal{L}_{iou}(t_i, \hat t_{\sigma(i)}) = 1 - \frac{ {\rm min}(t_i^e,\hat t_{\sigma(i)}^e) - {\rm max}(t_i^s,\hat t_{\sigma(i)}^s) }{ {\rm max}(t_i^e,\hat t_{\sigma(i)}^e) - {\rm min}(t_i^s,\hat t_{\sigma(i)}^s) }\,,
\label{eq:iou_loss}
\end{align}
where $t^s$ and $t^e$ denote the start and end timestamp, respectively. If two time segments $t_i$ and $\hat t_{\sigma(i)}$ perfectly overlap, the loss becomes 0; if they do not overlap at all, the loss becomes greater than 1.

\begin{figure}[t!]
    \setlength\tabcolsep{0.2em}
    \begin{center}
    \resizebox{\linewidth}{!}{
    \begin{tabular}{ccc}
        \multicolumn{3}{c}{Training curves} \\
        \multicolumn{3}{c}{\includegraphics[width=1.06\linewidth,height=0.29\textwidth]{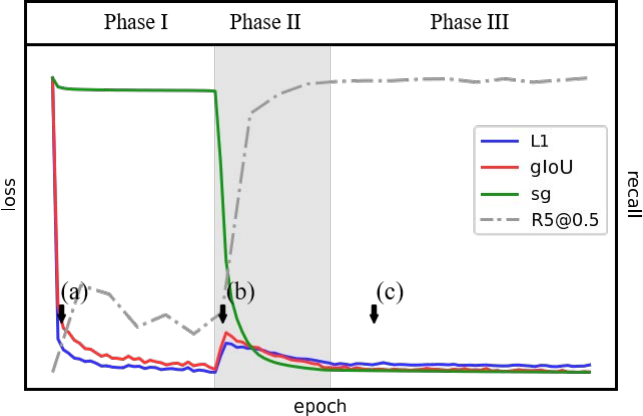}} \\
        \addlinespace[0.5em]
        \rot{\quad Video~\&~Queries} &
        \includegraphics[width=0.690\linewidth,height=0.15\textwidth]{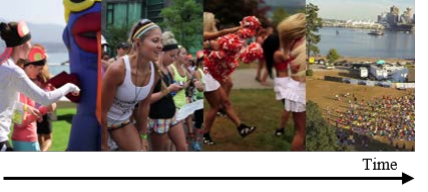} & \includegraphics[width=0.250\linewidth,height=0.15\textwidth]{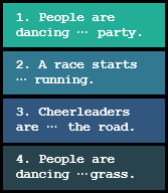} \\
        \rot{Ground truth} &
        \includegraphics[width=0.690\linewidth,height=0.1\textwidth]{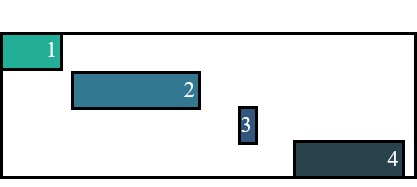} & \includegraphics[width=0.250\linewidth,height=0.1\textwidth]{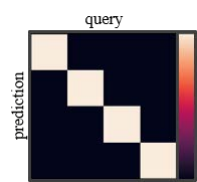} \vspace{-1em} \\
        \rot{\qquad (a)} &
        \subfigure{\includegraphics[width=0.690\linewidth,height=0.1\textwidth]{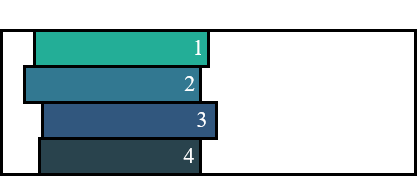}\label{fig:a_segments}} & \includegraphics[width=0.250\linewidth,height=0.1\textwidth]{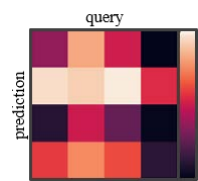} \vspace{-1em} \\
        \rot{\qquad (b)} &
        \subfigure{\includegraphics[width=0.690\linewidth,height=0.1\textwidth]{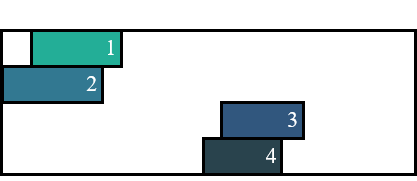}\label{fig:b_segments}} & \includegraphics[width=0.250\linewidth,height=0.1\textwidth]{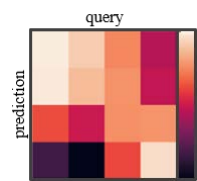} \vspace{-1em} \\
        \rot{\qquad (c)} &
        \subfigure{\includegraphics[width=0.690\linewidth,height=0.1\textwidth]{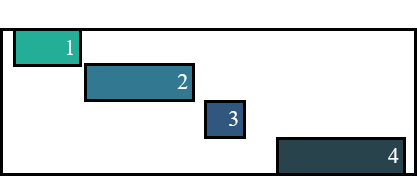}\label{fig:c_segments}} & \includegraphics[width=0.250\linewidth,height=0.1\textwidth]{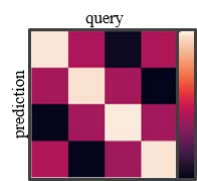} \\
         & Predicted time segments & Correspondence\\
    \end{tabular}
    }
    \end{center}
    \caption{
    Visualization of time segment predictions (left) and prediction-query correspondences (right) at three different points along the training curves (top):~\subref{fig:a_segments} At early training, neither segments nor order is accurate.~\subref{fig:b_segments} During the search space \textbf{exploration}, the segments are in the process of aligning with the targets, but they are unordered.~\subref{fig:c_segments} After proposals \textbf{match} the corresponding targets, the predicted segments are accurately aligned with the paired targets.
    }
    \label{fig:explore_and_match}
    \vspace{\belowfigcapmargin}
\end{figure}

\vspace{\paramargin}\paragraph{Final set prediction loss.}
The target query prediction and time segment prediction are factored into the matching cost.
We define matching cost using these notations:
\begin{align}
\mathcal{C}_{match}(y_i, \hat y_{\sigma(i)}) = &-\indic{q_i\neq\noobject}\hat p_{\sigma(i)}(q_i) \nonumber \\
&+ \indic{q_i\neq\noobject} \mathcal{L}_{loc}(t_{i}, \hat t_{\sigma(i)}) \,,
\label{eq:match_cost_2}
\end{align}
where $\mathds{1}$ indicates the indicator function. Here, we consider the $K$ matched predictions as positives (\ie, $q_i\neq\noobject$), and the remaining $(N-K)$ predictions as negatives (\ie, $q_i=\noobject$).
Contrary to the loss, we do not use the negative log-likelihood for the \sgl, but rather approximate it to $1-p_{\sigma(i)}(q_i)$.
We omit a constant $1$ since it does not change the matching.
Based on the matching results, our final set prediction loss is defined as:
\eqnsm{set_loss}{\mathcal{L}_{set}(y, \hat y) = \sum_{i=1}^N \left[\lambda_{\rm sg}\mathcal{L}_{sg}(q_i) + \indic{q_i\neq\noobject} \mathcal{L}_{loc}(t_{i}, \hat t_{\hat{\sigma}}(i))\right]\,,}
where $\lambda_{\rm sg}$ is a loss coefficient. Only the positives are optimized to predict the corresponding ground truth time segments.

\vspace{\paramargin}\paragraph{Unifying two streams of methods.}
Our approach inherits only the advantages from the \pb and the \pf methods.
We use the proposals, the core concept of the \pb methods, to encourage positive proposals to have higher similarities and suppress the negative proposals to have lower similarities with ground truth.
However, since \pb methods view NLVG as a classification problem, their performances are largely limited by hand-crafted components, such as predefined anchors and windows.
Our approach differs from the \pb methods in that it incorporates the flexibility of \pf methods.
We make every proposal learnable, allowing them to be fine-tuned within the training pipeline and dynamically transformed into more reliable proposals without the need for heuristics.

The combination of training ingredients condenses into a novel learning paradigm named \enm.
As shown in~\figref{fig:explore_and_match}, the \sgl and the \tll tend to show different patterns in training curves, where the former generates a cliff-like loss curve and the latter degrades smoothly.
At the beginning of the training~(Phase I:~\figref{fig:a_segments}), predicted time segments are almost a random initialization without order.
Before the sharp drop of a set guidance loss~(Phase II:~\figref{fig:b_segments}), a set of time segments aligns with a set of ground truths in a target-agnostic manner.
Interestingly, as the set guidance loss decreases, the $\ell_1$ loss and gIoU loss rebound slightly to reorganize predictions to be target-specific.
When all losses converge~(Phase III:~\figref{fig:c_segments}), time segments become accurate to match the target query. We empirically found that our method leads proposals to explore the search space, and then try to accurately match the target.
We note that the whole process is carried out in a systematic and holistic manner.

\begin{savenotes}
\begin{table*}[t!]
    \setlength{\tabcolsep}{0.2em}
        \centering
        \resizebox{\linewidth}{!}{
        \begin{tabular}{ccccccccc|ccccc}
            \toprule[0.2em]
            \multicolumn{4}{c}{}&\multicolumn{5}{c}{\textit{ActivityCaptions}}&\multicolumn{5}{c}{\textit{Charades-STA}}\\
            \cmidrule(l){5-9} \cmidrule(l){10-14}
            \multicolumn{3}{c}{Methods}&Venue&R1@0.5&R1@0.7&R5@0.5&R5@0.7&mIoU&R1@0.5&R1@0.7&R5@0.5&R5@0.7&mIoU\\
            \midrule[0.1em]
            \multicolumn{2}{c|}{\multirow{5}{*}{\rot{\tiny{proposal-based}}}}&\textsc{CTRL}~\cite{gao2017tall}&{\color{gray}ICCV2017}&-&-&-&-&-&23.63&8.89&58.92&29.52&-\\
            \multicolumn{2}{c|}{}&\textsc{TGN}~\cite{chen2018temporally}&{\color{gray}EMNLP2018}&27.93&-&44.20&-&-&-&-&-&-&-\\
            \multicolumn{2}{c|}{}&\textsc{2D-TAN}~\cite{zhang2020learning}&{\color{gray}AAAI2020}&44.51&26.54&77.13&61.96&-&39.70&27.1&80.32&51.26&-\\
            \multicolumn{2}{c|}{}&\textsc{CSMGAN}~\cite{liu2020jointly}&{\color{gray}ACMMM2020}&49.11&29.15&77.43&59.63&-&-&-&-&-&-\\
            \multicolumn{2}{c|}{}&\textsc{MSA}~\cite{zhang2021multi}&{\color{gray}CVPR2021}&48.02&31.78&78.02&63.18&-&-&-&-&-&-\\
            \midrule
            \multicolumn{2}{c|}{\multirow{5}{*}{\rot{\tiny{proposal-free}}}}&\textsc{ABLR}~\cite{yuan2019find}&{\color{gray}AAAI2019}&36.79&-&-&-&36.99&-&-&-&-&-\\
            \multicolumn{2}{c|}{}&\textsc{DEBUG}~\cite{lu2019debug}&{\color{gray}EMNLP2019}&39.72&-&-&-&39.51&37.39&17.69&-&-&36.34\\
            \multicolumn{2}{c|}{}&\textsc{DRN}~\cite{zeng2020dense}&{\color{gray}CVPR2020}&45.45&24.36&77.97&50.30&-&45.40&26.40&88.01&55.38&-\\
            \multicolumn{2}{c|}{}&\textsc{VSLNET}~\cite{zhang2020span}&{\color{gray}ACL2020}&43.22&26.16&-&-&43.19 &47.31&\textbf{30.19}&-&-&45.15\\
            \multicolumn{2}{c|}{}&\textsc{CPNET}~\cite{li2021proposal}&{\color{gray}AAAI2021}&40.65&21.63&-&-&40.65&40.32&22.47&-&-&37.36\\
            \midrule
            \multicolumn{2}{c|}{\multirow{2}{*}{\rot{etc}}}&\textsc{BPNET}~\cite{xiao2021boundary}&{\color{gray}AAAI2021}&42.07&24.69&-&-&42.11&38.25&20.51&-&-&38.03\\
            \multicolumn{2}{c|}{}&\textsc{CBLN}~\cite{liu2021context}&{\color{gray}CVPR2021}&48.12&27.60&\textbf{79.32}&\textbf{63.41}&-&47.94&28.22&88.20&\textbf{57.47}&-\\
            \midrule
            \multicolumn{4}{c}{\textbf{\lvtr-C3D} (Ours)}&53.27&27.93&78.19&57.82&51.00&47.15&25.72&86.91&53.19&44.26\\
            \rowcolor{Gray}
            \multicolumn{4}{c}{\textbf{\lvtr-CLIP} (Ours)}&\textbf{58.79}&\textbf{33.38}&77.47&59.68&\textbf{53.00}&\textbf{49.11}&26.59&\textbf{88.50}&55.99&\textbf{47.13}\\
            \bottomrule
        \end{tabular}
        }
    \vspace{\abovetabcapmargin}
    \caption{
        \textbf{Comparison with the state-of-the-arts} on two benchmark datasets (in the order of \textit{ActivityCaptions} and \textit{Charades-STA}).
    }
    \label{tab:sota}
    \vspace{\belowtabcapmargin}
\end{table*}
\end{savenotes}

\section{Experiments}
We first describe our experimental settings.
Next, we report our main results on two challenging benchmarks: \textit{ActivityCaptions}~\cite{caba2015activitynet, krishna2017dense} and \textit{Charades-STA}~\cite{gao2017tall}.
Lastly, we provide detailed ablation studies on the model variants and losses, and analyze how \lvtr works with visualizations.

\subsection{Experimental Setup}
\vspace{\paramargin}\paragraph{Datasets.}
1) \textbf{\textit{ActivityCaptions}}~\cite{caba2015activitynet, krishna2017dense} contains about 20K untrimmed videos with language descriptions and temporal annotations, which was originally developed for the task of dense video captioning~\cite{krishna2017dense}.
Following the convention, we used $val_1$ for validation and $val_2$ for testing since the test annotations are not publicly released. We also followed the standard split~\cite{yuan2019find}.
2) \textbf{\textit{Charades-STA}}~\cite{gao2017tall} is built on Charades~\cite{sigurdsson2016hollywood} and contains 6,672 videos of daily indoors activities. Each video is about 30 seconds long on average. We employed 12,408 video-sentence pairs for train and 3,720 pairs for test. 

\vspace{\paramargin}\paragraph{Evaluation metrics.}
Following~\cite{yuan2019find,lu2019debug}, we adopted two standard evaluation metrics for NLVG: 1) ``\textbf{R$\alpha$@$\mu$}", which denotes the percentage of test samples that have at least one correct result in top-$\alpha$ retrieved results, \ie, recall; here, the correct results indicate that IoU with ground truth is larger than threshold $\mu$. 2) ``\textbf{mIoU}”, which averages the IoU between predictions and ground truths over entire testing samples to compare the overall performance.

\vspace{\paramargin}\paragraph{Technical details.}
We trained \lvtr using AdamW~\cite{loshchilov2017decoupled} with an initial learning rate of 1e-4 and weight decay of 1e-4 for a batch size of 16. We used a linear learning rate decay by a factor of 10. We considered Xavier initialization~\cite{glorot2010understanding} to set the initial values of all transformer weights. We used 64 frames that are uniformly sampled from video with four sentences as an input. We resized every frame to 112$\times$112. The number of learnable proposals is proportionally set to 10 times the number of input queries.
For a fair evaluation with baselines, we extracted video representations with C3D~\cite{tran2015learning} pretrained on Sports-1M~\cite{karpathy2014large}, and for the language part, we initialized each word with GloVe embeddings~\cite{pennington2014glove} and obtained sentence representation via 2-layer bi-LSTM~\cite{hochreiter1997long}. In training, we set our loss weight $\lambda_{\rm L1}:\lambda_{\rm iou}:\lambda_{\rm sg}$ to $1:3:2$. We also used an auxiliary decoding loss~\cite{al2019character} in decoder layers to speed up the convergence. The initial proposals are filled with learnable weights~\cite{carion2020end}.

\subsection{Main Results}
\vspace{\paramargin}\paragraph{Comparison with state-of-the-art approaches.}
We compared \lvtr against recently proposed NLVG methods, which can be largely categorized into three groups: 1) \pb: CTRL~\cite{gao2017tall}, TGN~\cite{chen2018temporally}, 2D-TAN~\cite{zhang2020learning}, CSMGAN~\cite{liu2020jointly}, MSA~\cite{zhang2021multi}, 2) \pf: ABLR~\cite{yuan2019find}, DEBUG~\cite{lu2019debug}, DRN~\cite{zeng2020dense}, VSLNET~\cite{zhang2020span}, CPNET~\cite{li2021proposal}, and
3) \textit{etc}: BPNet~\cite{xiao2021boundary}, CBLN~\cite{liu2021context}.
\lvtr with C3D backbone (\lvtr-C3D) sets new state-of-the-arts on two benchmarks (see~\tabref{tab:sota}): \textit{ActivityCaptions}~\cite{caba2015activitynet,krishna2017dense} and \textit{Charades-STA}~\cite{gao2017tall}.
Especially for R1@0.5 metric on \textit{ActivityCaptions} dataset, \lvtr-C3D achieved about 10\% performance gain compared to CBLN~\cite{liu2021context}.
We further improved the performance of \lvtr by using CLIP~\cite{radford2021learning} as a backbone (\lvtr-CLIP) where a massive amount of image-text pairs are pre-trained with contrastive learning.
Even freezing the backbone in the training phase, we observed that CLIP significantly boosts the performance, implying that visual-linguistic domain alignment is important.

\vspace{\paramargin}\paragraph{Inference speed.}
We compared several methods in terms of inference speed required to localize a single sentence query in~\figref{fig:teaser}.
Our \lvtr takes an average of 10ms to process a language query on \textit{ActivityCaptions} dataset.
\lvtr runs much faster than the previous NLVG methods, especially 2$\times$ faster than DEBUG~\cite{lu2019debug}.
Furthermore, our set matching formulation eliminates the time-consuming pre-processing or post-processing stage, such as dense proposal generation and non-maximum suppression.

\begin{figure}[t!]
    \setlength\tabcolsep{0.2em}
    \begin{center}
    \resizebox{\linewidth}{!}{
    \begin{tabular}{c|c}
        \rot{\quad \textbf{Video}} &
        \includegraphics[width=0.98\linewidth]{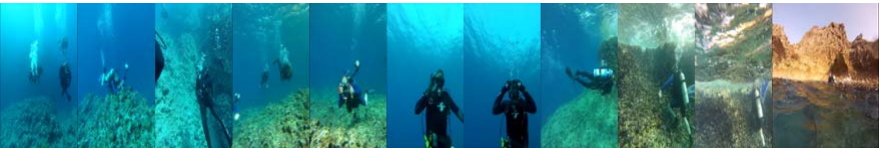} \\
        \midrule
        \rot{\quad \textbf{Language queries}} &
        \includegraphics[width=0.98\linewidth]{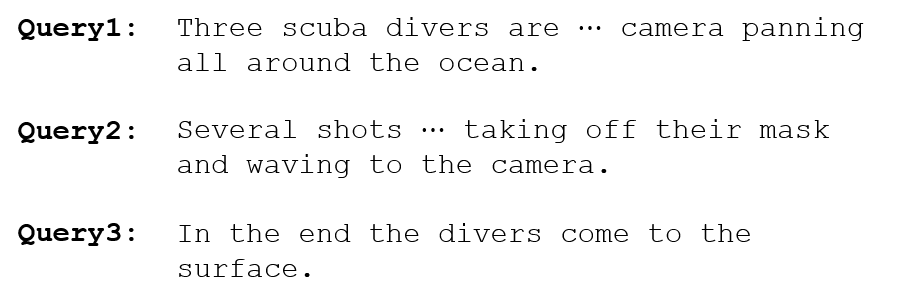} \\
        \midrule
        \rot{\qquad \textbf{GT}} &
        \includegraphics[width=0.98\linewidth]{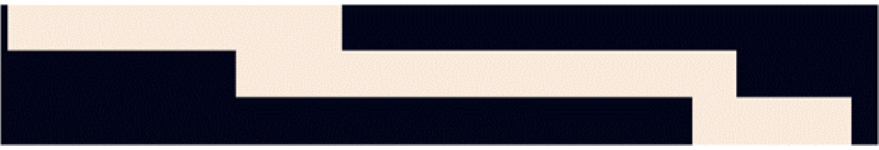} \\
        \midrule
        \rot{\qquad\qquad \textbf{Time segment predictions}} &
        \includegraphics[width=0.98\linewidth]{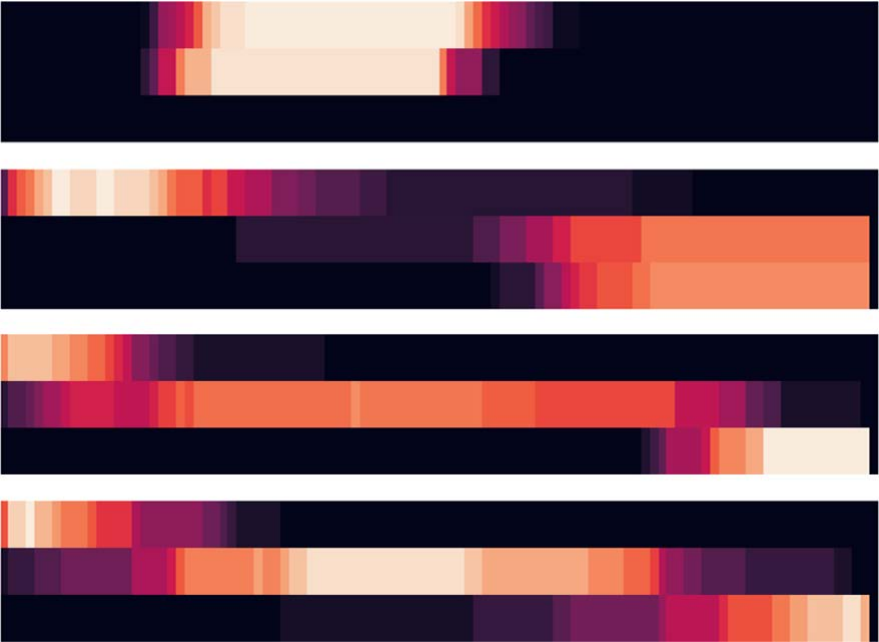} \\
    \end{tabular}
    }
    \end{center}
    \caption{
    \textbf{Visualization of time segment predictions throughout training under \enm scheme.}
    The four predictions are in time order from top to bottom, where the first two of them show the \textit{explore} process and the last two show the \textit{match} process.
    The brighter the color, the more time segments predicted by proposals overlap.
    }
    \label{fig:training}
\end{figure}

\subsection{Analysis}
\vspace{\paramargin}\paragraph{Training with Explore-And-Match scheme.}
In~\figref{fig:training}, we investigated the behavior of \lvtr during training under the \enm scheme. 
Beginning with the random initialization, the proposals start to generate some variations in their predictions (1st).
Thereafter, they become a state in which they can adapt to any time segment by slightly overlapping the boundaries of two ground truths (2nd).
Then, each of their identities is determined, and their time segments are adjusted accordingly (3rd).
After matching the identity, the proposals properly fit the relevant time segment in a fine-grained manner (4th).
To our surprise, despite all training losses are given at once, the training process follows the form of a divide-and-conquer-like approach.
We hypothesize that a carefully designed training scheme facilitates this systematic behavior.

\begin{table}[t!]
    \setlength{\tabcolsep}{0.3em}
    \begin{center}
    \resizebox{\linewidth}{!}{
    \begin{tabular}{@{}ccc|cccccc@{}}
        \toprule[0.2em]
        \textbf{$\mathcal{L}_{sg}$} & \textbf{$\mathcal{L}_{\rm L1}$} & \textbf{$\mathcal{L}_{iou}$}    &{R1@0.5}&{R1@0.7}&{R5@0.5}&{R5@0.7}& {mIoU}\\ \midrule
        \cmark&      &      &  24.30 & 9.60 & 24.69 & 9.75 & 26.9 \\
        \cmark&\cmark&      &  41.42 & 18.25 & 72.55 & 54.90 & 41.25 \\
        \cmark&      &\cmark&  50.15 & 27.90 & 72.32 & 55.50 & 48.01 \\
        \rowcolor{Gray}
        \cmark&\cmark&\cmark&  \textbf{58.79} & \textbf{33.38} & \textbf{77.47} & \textbf{59.68} & \textbf{53.00} \\
        \bottomrule
    \end{tabular}
    }
    \vspace{\abovetabcapmargin}
    \caption{
        \textbf{\textbf{Ablation results of the loss functions.}}
    }
    \label{tab:loss_ablations}
    \vspace{\belowtabcapmargin}
    \end{center}
\end{table}

\vspace{\paramargin}\paragraph{Loss ablations.}
We analyzed the impact of the loss terms in~\tabref{tab:loss_ablations}: $\ell_1$ loss ($\mathcal{L}_{\rm L1}$), gIoU loss ($\mathcal{L}_{iou}$), and set guidance loss ($\mathcal{L}_{sg}$).
Since matching the target query is essential, we always used \sgl for all cases.
When both L1 and gIoU are disabled, the predictions are collapsed; thus, R1@0.5 and R5@0.5 showed almost the same results.
When either L1 loss or gIoU loss is disabled, performance suffered significantly, implying that they are both required for accurate temporal localization.
As using all three losses yielded the best result, we confirmed that two sub-losses of \tll (L1 and gIoU) operate complementarily with absolute or relative criteria for time segment prediction.

\begin{table}[t!]
    \setlength{\tabcolsep}{0.3em}
    \begin{center}
    \resizebox{\linewidth}{!}{
    \begin{tabular}{@{}c|cccccc@{}}
        \toprule[0.2em]
        Methods & R1@0.5 & R1@0.7 & R5@0.5 & R5@0.7 & mIoU\\
        \midrule
        Sim  &13.11 & 2.75 & 13.15 & 2.79 & 23.73\\
        Att  & 34.36 & 18.10 & \textbf{82.42} & \textbf{63.31} & 39.16\\
        \rowcolor{Gray}
        \textbf{Cos} &  \textbf{58.79} & \textbf{33.38} & 77.47 & 59.68 & \textbf{53.00}\\
        \bottomrule
    \end{tabular}
    }
    \vspace{\abovetabcapmargin}
    \caption{
        \textbf{Choices for pred-query correspondence measure.}
    }
    \label{tab:query_order_pred}
    \vspace{\belowtabcapmargin}
    \end{center}
\end{table}

\vspace{\paramargin}\paragraph{Correspondence measures.}
We compared the various measures to calculate the correspondence between prediction and query in~\tabref{tab:query_order_pred}, which is then used in \sgl.
In practice, we considered proposal-target matching using decoder output and the textual part of encoder output.
The encoder-decoder attention weight (Att) is an intuitive way of determining which part of the encoder output each proposal corresponds to.
Since it has direct access to the global context, it performed well especially for the R5 metric, but fells short for the rigorous R1 metric.
We observed that using cosine similarity (Cos) dramatically improves performance than directly applying dot product similarity (Sim), meaning that removing the size constraint eases optimization.

\begin{table}[t!]
    \setlength{\tabcolsep}{0.3em}
    \begin{center}
    \resizebox{\linewidth}{!}{
    \begin{tabular}{@{}cc|ccccc@{}}
        \toprule[0.2em]
        vid & txt & {R1@0.5}  & {R1@0.7} & {R5@0.5} & {R5@0.7} & {mIoU}   \\ \midrule
               &        & 25.71 & 12.69 & 66.34 & 41.45 & 29.85 \\
        \cmark &        & 22.86 & 10.75 & 63.95 & 40.83 & 30.11 \\
               & \cmark & 38.18 & 16.05 & 74.56 & 54.58 & 40.88 \\
        \rowcolor{Gray}
        \cmark & \cmark & \textbf{58.79} & \textbf{33.38} & 77.47 & 59.68 & \textbf{53.00} \\
        \bottomrule
    \end{tabular}
    }
    \vspace{\abovetabcapmargin}
    \caption{
        \textbf{\textbf{Positional Encodings.}}
    }
    \label{tab:pos_encoding}
    \vspace{\belowtabcapmargin}
    \end{center}
    \vspace{-4mm}
\end{table}

\vspace{\paramargin}\paragraph{Positional encodings.}
In~\tabref{tab:pos_encoding}, we ablated the positional encodings of \lvtr.
First, we disabled positional encoding for both video and text input.
As expected, temporally unorganized input severely degrades performance. 
The positional encoding of each modality input is then removed in turn.
When the video positional encodings are disabled, the model can no longer utilize temporally coordinated video contexts.
Also, the temporal clue provided by textual positional encoding is significant in textual input since it aids in organizing the order of events.
We used both positional encodings since both positional encoding largely contributes to the performance.
To align the video and text in a different time axis, we employed separate positional encodings for each modality input.

\begin{table}[t!]
    \setlength{\tabcolsep}{0.3em}
    \begin{center}
    \resizebox{\linewidth}{!}{
    \begin{tabular}{@{}cc|cccccc@{}}
        \toprule[0.2em]
        \#enc & \#dec & {R1@0.5}  & {R1@0.7} & {R5@0.5} & {R5@0.7} & {mIoU}\\ \midrule
        1 & 1 & 48.16 & 25.55 & \textbf{79.38} & \textbf{64.34} & 48.02 \\
        2 & 1 & 48.20 & 26.08 & 77.57 & 64.05 & 47.67 \\
        1 & 2 & 48.30 & 25.40 & 75.39 & 57.72 & 47.97 \\ 
        2 & 2 &  55.22 & 31.13 & 76.39 & 61.65 & 50.99 \\
        3 & 3 &  53.32 & 26.62 & 74.65 & 58.44 & 48.92 \\ 
        \rowcolor{Gray}
        \textbf{4} & \textbf{4} &  \textbf{58.79} & 33.38 & 77.47 & 59.68 & \textbf{53.00} \\
        5 & 5 &  56.11 & \textbf{33.82} & 79.15 & 60.70 & 52.04 \\
        \bottomrule
    \end{tabular}
    }
    \vspace{\abovetabcapmargin}
    \caption{
        \textbf{\textbf{Model variants w.r.t encoder-decoder size.}}
    }
    \label{tab:num_layers}
    \vspace{\belowtabcapmargin}
    \end{center}
\end{table}

\vspace{\paramargin}\paragraph{Model size.}
To examine the effect of model size, we varied the number of encoder-decoder layers (see~\tabref{tab:num_layers}). We first compared the two asymmetric structures (\#Enc-\#Dec): 2-1 \textit{vs.} 1-2. Compared to the former, the latter fells 2.18 points in R5@0.5 and 6.33 points in R5@0.7 metrics, showing that the contextualization in the encoder is important in generating high-quality proposals.
As the size of the transformer increases, the R1 metric gradually improves, while R5 does not change appreciably.
This suggests that increasing the size of the transformer has the effect of focusing on selecting better predictions among the candidates.
However, considering that the performance degraded in 5-5, stacking more encoder-decoders does not always guarantee higher performance.
Among several variants, we found that 4-4 shows the optimal performance.

\begin{table}[t!]
    \setlength{\tabcolsep}{0.3em}
    \begin{center}
    \resizebox{\linewidth}{!}{
    \begin{tabular}{@{}c|cccccc@{}}
        \toprule[0.2em]
        \#proposals & {R1@0.5}  & {R1@0.7} & {R5@0.5} & {R5@0.7} & {mIoU} \\ \midrule
        5 &  48.74  & 25.40 & \textbf{86.37}  & \textbf{70.68} & 48.03 \\
        \rowcolor{Gray}
        \textbf{10} & \textbf{58.79} & \textbf{33.38} & 77.47 & 59.68 & \textbf{53.00} \\
        15 & 34.26 & 16.17 & 65.59 & 47.35 & 39.74 \\
        20 & 6.64 & 2.16 & 18.02 & 6.11 & 13.07 \\ \bottomrule
    \end{tabular}
    }
    \vspace{\abovetabcapmargin}
    \caption{
        \textbf{\textbf{Number of learnable proposals per query.}}
    }
    \label{tab:num_proposals}
    \vspace{\belowtabcapmargin}
    \end{center}
\end{table}

\vspace{\paramargin}\paragraph{Number of learnable proposals.}
We searched for the optimal number of proposals per language query in~\tabref{tab:num_proposals}.
A small number of proposals limits sufficient interactions between positives and negatives, resulting in sub-optimal performance, whereas an excessive quantity of proposals reduces accuracy by generating too many negatives. There is a trade-off between R5 and mIoU metrics around the appropriate number of proposals. Between them, having 10 learnable proposals per query yielded the best results.

\begin{figure}[t!]
    \centering
    \small
    \hspace{-1.3em}
    \includegraphics[width=1.04\linewidth]{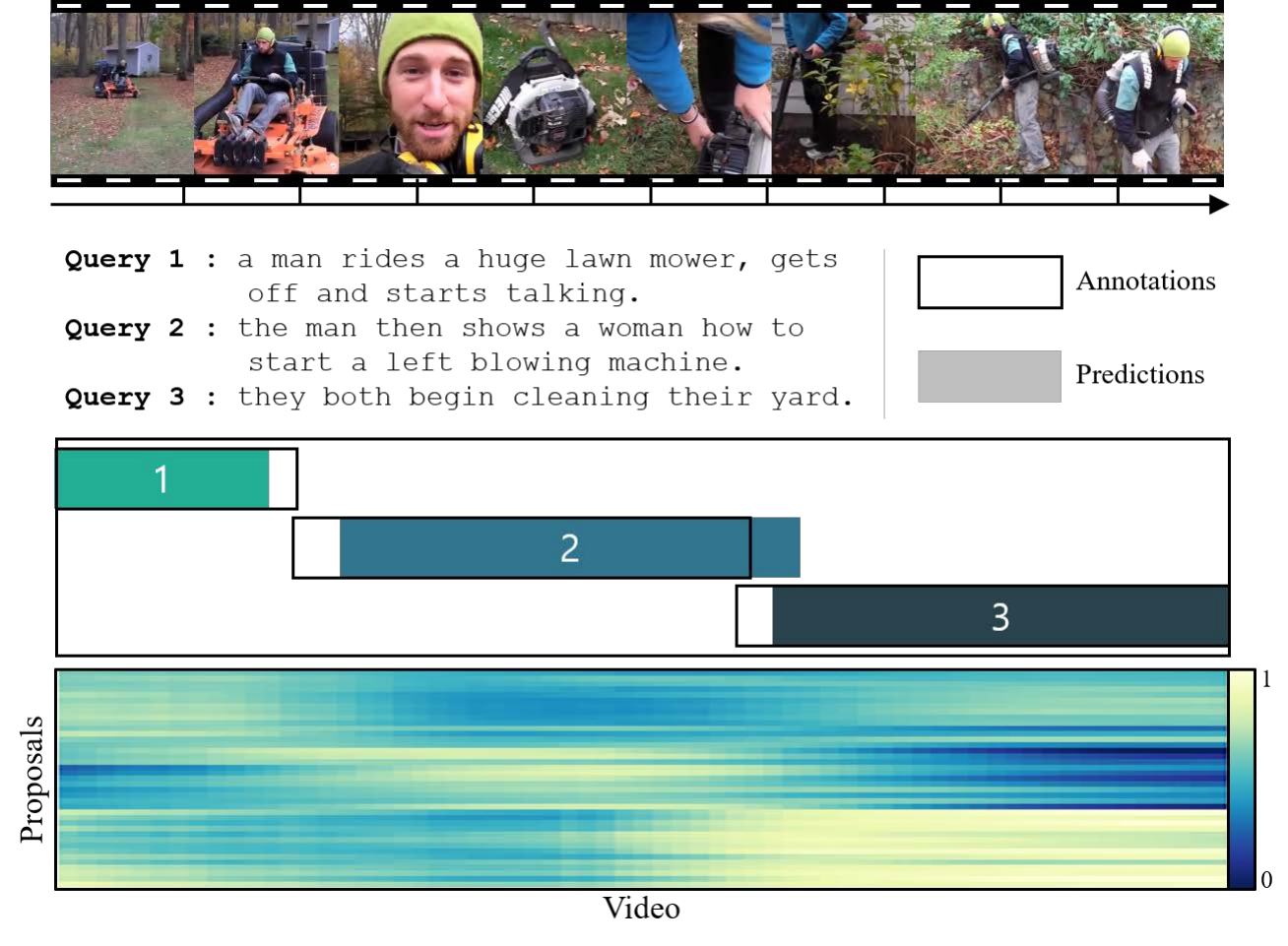}
    \caption{\textbf{NLVG results (middle) of \lvtr with proposal-video attention map (bottom).}
    The bottom color map indicates the attention to the video time segment (column) of each proposal (row), where the brighter the color, the higher attention is.
    Note that each subset of learnable proposals attends to the corresponding video contexts to predict the target time segments.
    }
    \label{fig:attention}
\end{figure}

\vspace{\paramargin}\paragraph{Attention visualization on qualitative example.}
In~\figref{fig:attention}, we show a sample NLVG result, where the bars lie along the time axis represent the time segments grounded on the query.
The predictions (color bars) generated by \lvtr nearly matched the target time segments (empty bars).
As shown in the proposal-video attention map (bottom), the time segments in which each subset of learnable proposals attend to are mostly overlapped with their corresponding time segment prediction; for example, the third subset attend to the end part of the video.
This implies that proposals within the same subset consider similar parts of the video contexts when predicting the target query.

\begin{figure}[t!]
    \centering
    \small
    \includegraphics[width=1.0\columnwidth]{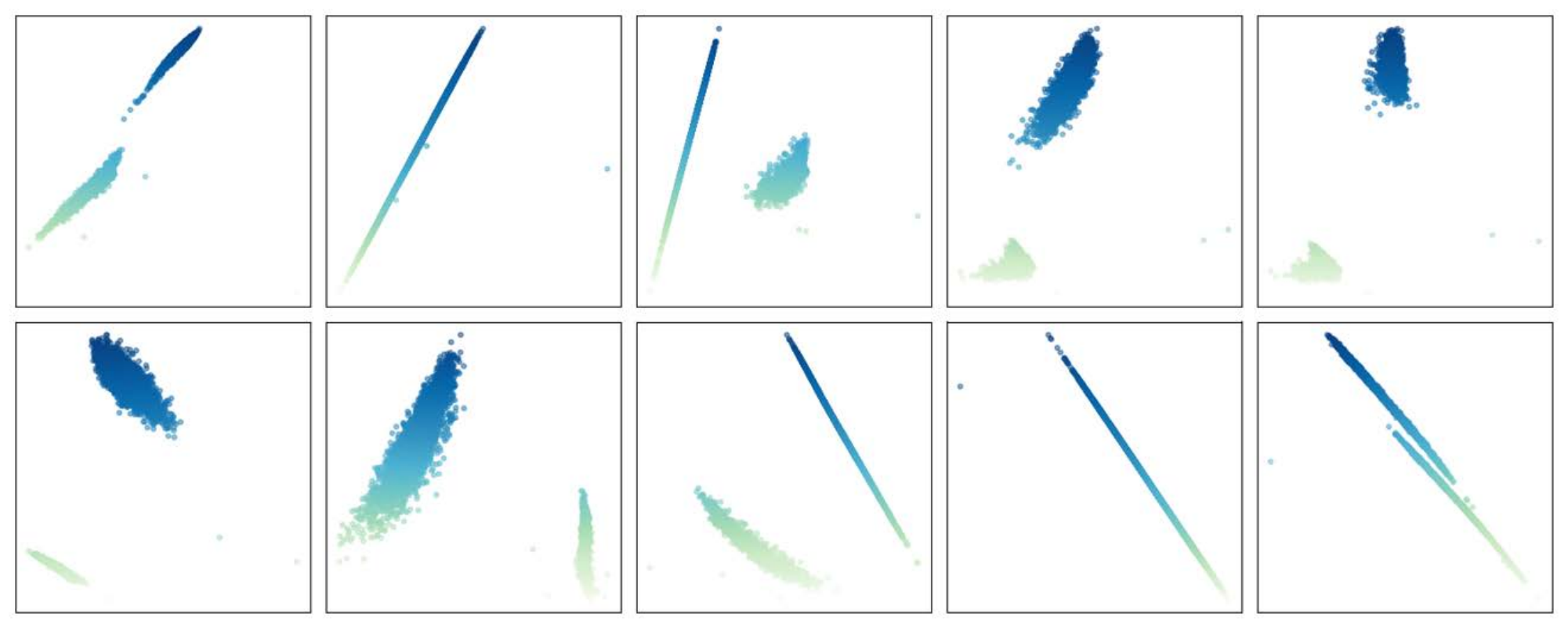}
    \caption{\textbf{Visualization of predicted time segments} on \textit{ActivityCaptions} for 10 out of all learnable proposals. Each prediction is represented by a colored point on the horizontal (center) and vertical (width) axes, where the color indicates the width. We observe that each learnable proposal learns to specialize on certain time zones and durations.
    }
    \label{fig:pred_dist}
\end{figure}

\vspace{\paramargin}\paragraph{Distribution of learnable proposals.}
We visualize the time segment predictions of 10 out of all learnable proposals in~\figref{fig:pred_dist}.
We observed that they exihibited  a variety of distinct patterns, implying that \lvtr learns unique specializations for each proposal.
More specifically, each proposal includes several operating modes attending to different time zones and durations.
For example, the top-third proposal learned about a long period of time at the beginning of the videos.
Overall, all proposals have a mode that predicts video-wide durations, denoted by the color blue.

\begin{table}[t!]
    \setlength{\tabcolsep}{0.3em}
    \begin{center}
    \resizebox{\linewidth}{!}{
    \begin{tabular}{@{}c|ccccc@{}}
        \toprule[0.2em]
        \textbf{$\lambda_{\rm L1}$:$\lambda_{\rm iou}$:$\lambda_{\rm sg}$} & {R1@0.5}  & {R1@0.7} & {R5@0.5} & {R5@0.7} & {mIoU}   \\ \midrule
        1:1:1 & 53.39 & 30.79 & \textbf{80.68} & 60.76 & 51.24 \\
        2:1:1 & 56.42 & 31.59 & 78.90 & \textbf{63.03} & 52.77 \\
        1:2:1 & 58.03 & 31.23 & 78.02 & 60.49 & 52.15 \\
        1:1:2 & 57.41 & 32.91 & 77.09 & 57.49 & \textbf{53.50} \\
        1:3:1 & 49.59 & 28.03 & 66.81 & 48.78 & 48.18 \\
        \rowcolor{Gray}
        1:3:2 & \textbf{58.79} & \textbf{33.38} & 77.47 & 59.68 & 53.00 \\
        \bottomrule
    \end{tabular}
    }
    \vspace{\abovetabcapmargin}
    \caption{
        \textbf{\textbf{Loss balancing parameters.}}
    }
    \label{tab:loss_hyperparams}
    \vspace{\belowtabcapmargin}
    \end{center}
\end{table}

\vspace{\paramargin}\paragraph{Loss hyperparameters.}
We searched for optimal loss hyperparameters in~\tabref{tab:loss_hyperparams}.
We begun by setting the loss coefficients to 1:1:1 by default. 
While set guidance loss ($\lambda_{\rm sg}$) is essential for query identity matching, the span localization loss ($\lambda_{\rm L1}$ and $\lambda_{\rm iou}$) directly affects the accurate video grounding.
This can be confirmed by varying the coefficient for each term to 2 one-by-one.
Among the three variations, we found that gIoU loss ($\lambda_{\rm iou}$) is the most important term in the loss function.
This is because the relative measure is more robust to varying spans shifted over various time distributions.
While maintaining the gIoU loss to hold the major term, 1:3:2 yielded the best results in our setting.

\begin{table}[t!]
    \setlength{\tabcolsep}{0.3em}
    \begin{center}
    \resizebox{\linewidth}{!}{
    \begin{tabular}{@{}c|ccccc@{}}
        \toprule[0.2em]
        \textbf{\#Frames} & {R1@0.5}  & {R1@0.7} & {R5@0.5} & {R5@0.7} & {mIoU}   \\ \midrule
        16 & 47.93 & 23.34 & 72.30 & 51.31 & 42.53 \\
        32 & 52.15 & 28.77 & 74.01 & 55.13 & 50.46 \\
        \rowcolor{Gray}
        \textbf{64} & \textbf{58.79} & \textbf{33.38} & \textbf{77.47} & 59.68 & \textbf{53.00} \\
        128 & 53.73 & 29.55 & 77.42 & \textbf{60.77} & 51.11 \\ 
        256 & 48.35 & 24.39 & 73.36 & 54.23 & 47.89 \\ \bottomrule
    \end{tabular}
    }
    \vspace{\abovetabcapmargin}
    \caption{
        \textbf{\textbf{Effect of the number of input frames.}}
    }
    \label{tab:num_frames}
    \vspace{\belowtabcapmargin}
    \end{center}
\end{table}

\begin{table}[t!]
    \setlength{\tabcolsep}{0.3em}
    \begin{center}
    \resizebox{\linewidth}{!}{
    \begin{tabular}{@{}c|ccccc@{}}
        \toprule[0.2em]
        \textbf{\#Sentences} & {R1@0.5}  & {R1@0.7} & {R5@0.5} & {R5@0.7} & {mIoU}   \\ \midrule
        2 & 33.86  & 17.31  & 70.41  & 44.00 & 38.73 \\
        3 & 46.15 & 22.03 & \textbf{81.58} & \textbf{63.17} & 47.05 \\ 
        \rowcolor{Gray}
        \textbf{4} &  \textbf{58.79} & \textbf{33.38} & 77.47 & 59.68 & \textbf{53.00} \\
        5 & 35.41 & 17.51& 69.19 & 48.74 & 39.58 \\ 
        \bottomrule
    \end{tabular}
    }
    \vspace{\abovetabcapmargin}
    \caption{
        \textbf{\textbf{Effect of the number of input sentences.}}
    }
    \label{tab:num_queries}
    \vspace{\belowtabcapmargin}
    \vspace{-3mm}
    \end{center}
\end{table}

\vspace{\paramargin}\paragraph{Input analysis.}
In order to examine the effect according to the number of input video frames and the number of input sentences, we varied the numbers in~\tabref{tab:num_frames} and~\tabref{tab:num_queries}, respectively.
As we expect more frames to bring more temporal knowledge, too few frames miss the exact moment when the event occurs, leading to decrease in performance.
However, the results reveal that a large number of frames does not always guarantees better results.
This implies that adding more frames cause a trade-off in the optimization while increasing the sequence length.
We found that 64 produces the best results.
Using multiple sentences as input queries allows us to take advantage of the temporal contexts between language queries.
In the R1 metric, using 4 sentences as an input outperforms using 3 sentences, while using 3 sentences as an input shows better results in the R5 metric.
This is due to the fact that the average number of existing sentences in training split of \textit{ActivityCaptions} is $3.739$.
We adopt 4 sentences as an input since we require a more accurate model on a stricter metric.

\begin{figure*}[pt!]
    \centering
    \setlength\tabcolsep{2.12345pt}
    \resizebox*{\linewidth}{!}{%
    \begin{tabular}{c|c}
    \toprule[0.2em]
        \color{blue}{\textbf{Success Case}} & \color{red}{\textbf{Failure Case}} \\
        \midrule[0.07em] \addlinespace[0.5em]
        \includegraphics[width=0.49\linewidth]{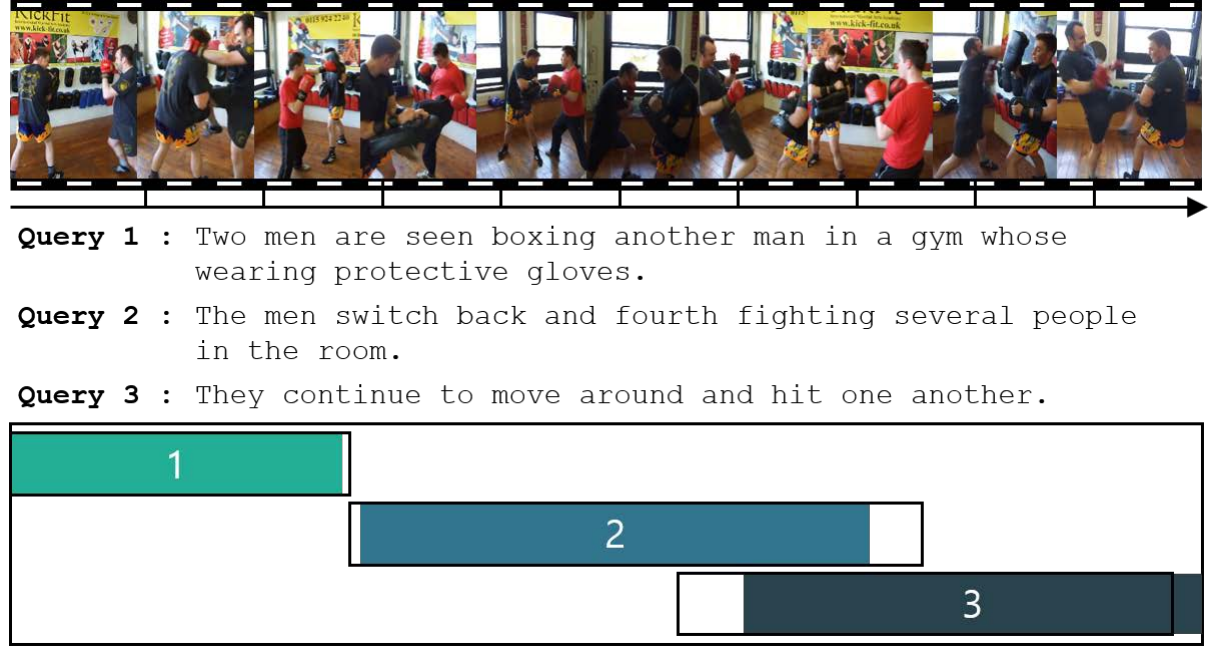} & \includegraphics[width=0.49\linewidth]{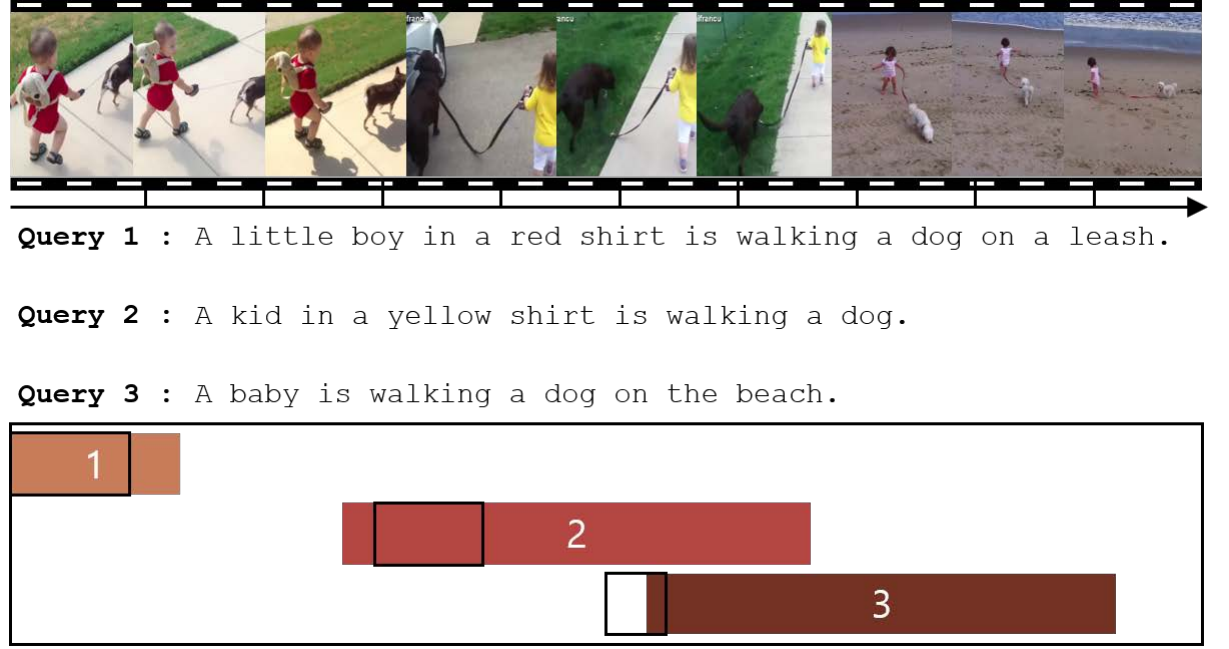} \\
        \includegraphics[width=0.49\linewidth]{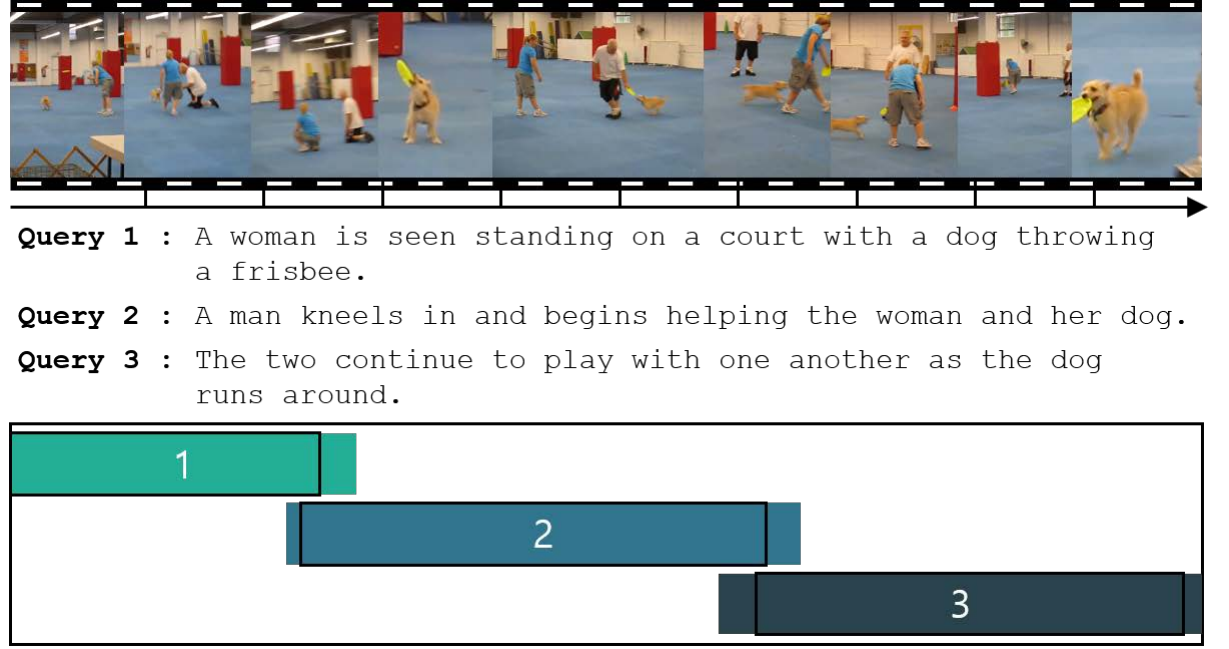} & \includegraphics[width=0.49\linewidth]{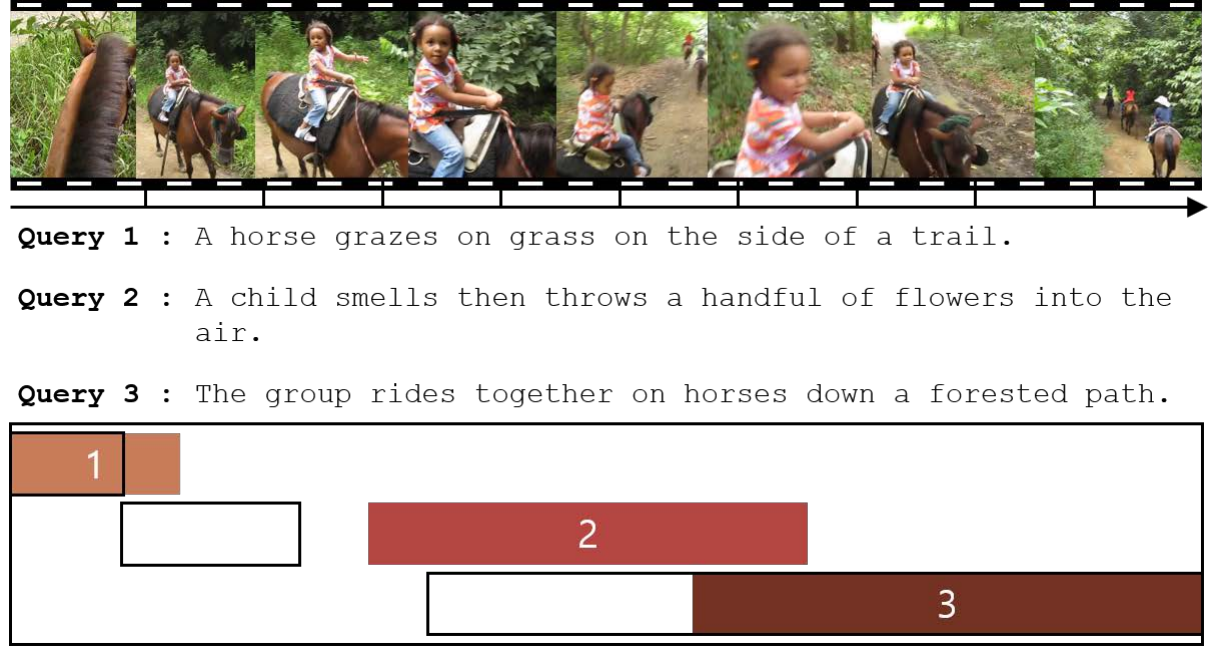} \\
        \includegraphics[width=0.49\linewidth]{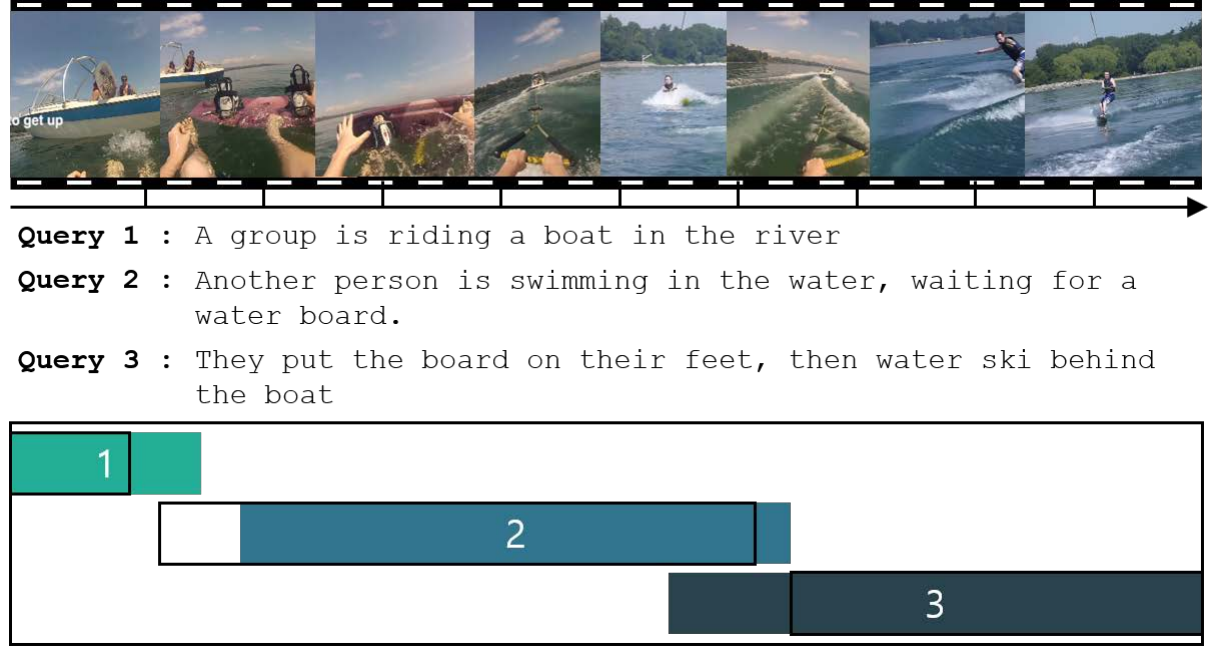} & \includegraphics[width=0.49\linewidth]{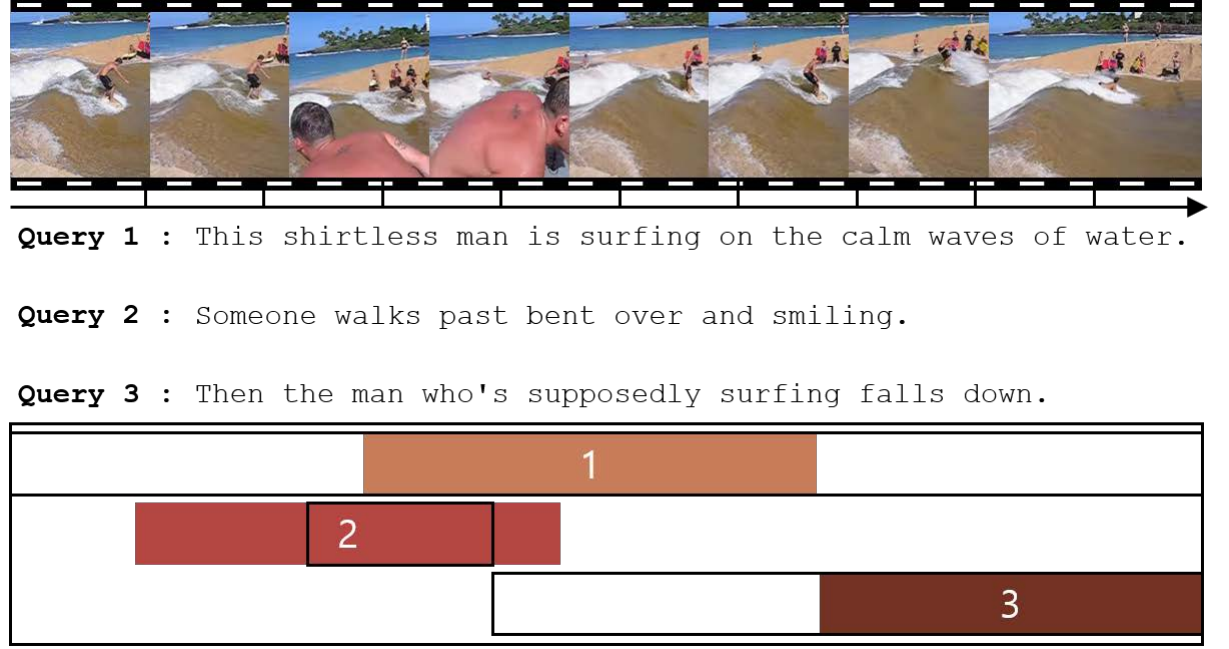} \\
        \includegraphics[width=0.49\linewidth]{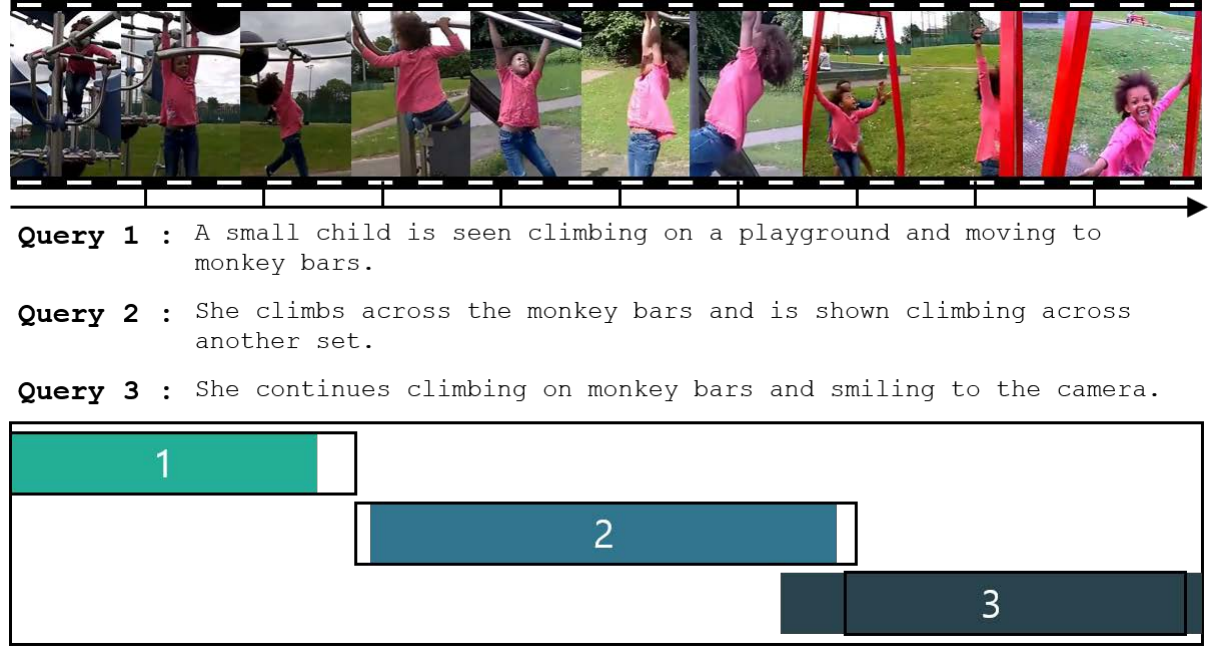} & \includegraphics[width=0.49\linewidth]{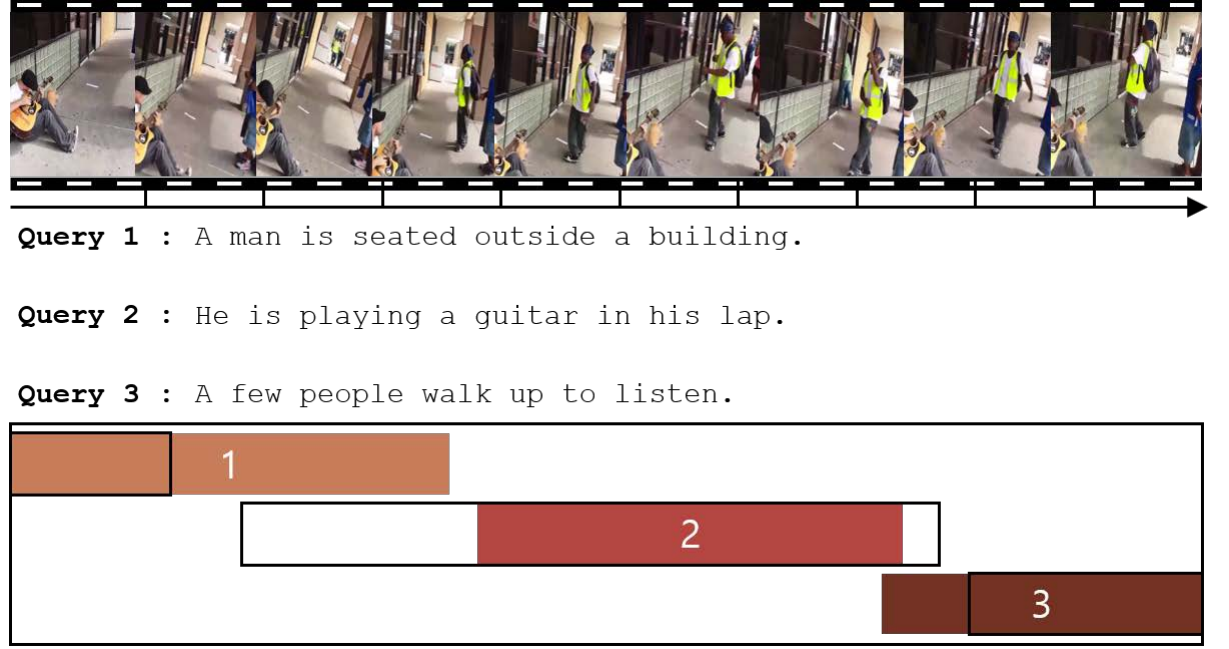} \\
        \bottomrule[0.1em]
    \end{tabular}%
    }
    \vspace{5pt}
    \caption{\textbf{Qualitative examples of success and failure cases} of \lvtr on the \textit{ActivityCaptions} dataset. The predicted time segment is considered correct only if it has sufficiently high IoU (\ie, IoU $>$ 0.5) with ground truth time segment. Empty bars represent ground truths, and colored bars represent predictions.}
    \label{fig:qualitative}
\end{figure*}

\subsection{Qualitative Results}
To better see how \lvtr understands the video contexts, we provide qualitative results and contrast the success and failure cases in~\figref{fig:qualitative}.
The results show that the \lvtr successfully identifies the object described in the query and accurately localize the time segment, even if multiple objects appear in the video (row 1\&2).
Moreover, \lvtr correctly reasons about the action that takes place from the first person point of view (row 3).
Lastly, even if the same object appears repeatedly, \lvtr distinguishes subtle contextual differences between them well (row 4).
However, \lvtr often fails to capture short-term events, especially when the object is too small (row 1\&2).
\lvtr suffers when the time the event takes place is too long (\eg, whole video length) (row 3).
Also, \lvtr fails when the labeled time segment and the actual time segment where the query description matches the video content are significantly different (row 4).

\section{Discussion}
Our framework inherits the popular DETR~\cite{carion2020end} framework that is for object detection, making it easy for practitioners to implement, yet we show that our framework can effectively solve the NLVG problem with little modifications. The reason for using the DETR framework is that if object detection is a problem of finding bounding boxes on the spatial axis, NLVG is a problem of finding bounding boxes on the temporal axis.

We designed the NLVG problem as a set prediction problem, which allowed us to integrate \textit{explore} and \textit{match} into a single step. For set prediction, we use learnable proposals, which should be able to generate flexible proposals as in proposal-free, while resolving the largest issue of proposal-based methods: redundant pre-generated proposals. To this end, the ideal method for learning the learnable proposals is to use the property of the Transformer, which models the pairwise interaction between all tokens in the input sequence. As we input the learnable proposals as a sequence to the Transformer decoder, each learnable proposal adjusts the time segments in consideration of other learnable proposals such that they are neither biased nor overlapped (as shown in~\figref{fig:pred_dist}).

Unlike typical NLVG methods, our framework can handle multiple queries (ofcourse single query too) --- our language encoder processes sentence units rather than word units, making multi-query learning possible.
With a fixed number of learnable proposals, our method can simultaneously predict multiple answers.
This is effective in that learnable proposals can utilize the temporal order between sentences, and it is efficient in that learnable proposals can predict multiple sentences at the same time (see the results in~\figref{fig:teaser}).
Our newly introduced \sgl matches the learnable proposals with multiple queries.
The \sgl divides the full set of learnable proposals into several subsets according to the number of input queries and induces each subset to be learned according to each query.
Combining the \sgl with \tll, our framework falls into the \enm scheme --— every learnable proposal first explores the search space, and then accurately matches the target.
Surprisingly, since our network is learned end-to-end, optimization for both losses occurs simultaneously, yet we can observe that all learnable proposals divide-and-conquer the problem (explore first and match next) holistically and systematically (see~\figref{fig:explore_and_match} and~\figref{fig:training}).

\section{Conclusion}
We have introduced \textit{\enm}, a new NLVG paradigm that unifies \pb and \pf approaches; our approach inherits the former concept while proposals are flexible as in the latter.
We viewed NLVG as a direct set prediction problem, and designed a transformer-based \lvtr to solve this problem.
\lvtr is end-to-end trainable and can predict time segments in parallel by utilizing abundant video-text contexts.
We employed bipartite matching in tandem with two key losses: 1) \sgl forces to match the target, and 2) \tll regresses each proposal to fit the corresponding time segment.
Our approach diversifies proposals in the \textit{explore} stage, and aligns each learnable proposal with specific target in the \textit{match} stage.
\lvtr achieved new state-of-the-art results on two challenging benchmarks (ActivityCaptions and Charades-STA) while doubling the inference speed.
We hope our exploration and findings facilitate future research on NLVG.

{\small
\bibliographystyle{ieee_fullname}
\bibliography{egbib}
}

\end{document}